\documentclass{egpubl_arxiv}
 
\JournalSubmission    
\usepackage[T1]{fontenc}
\usepackage{dfadobe}  

\usepackage{cite}  
\BibtexOrBiblatex

\electronicVersion
\PrintedOrElectronic
\ifpdf \usepackage[pdftex]{graphicx} \pdfcompresslevel=9
\else \usepackage[dvips]{graphicx} \fi

\usepackage{egweblnk} 

\usepackage{bm}
\usepackage{xspace}
\usepackage{graphicx}
\usepackage{booktabs}
\usepackage{multirow}
\usepackage{color}
\usepackage{colortbl}
\usepackage{amssymb}
\usepackage{amsmath}
\usepackage{cite}

\usepackage{makecell}

\usepackage{caption}

\newcommand{\xmark}{\ding{55}}

\usepackage[table]{xcolor}
\definecolor{mygray}{gray}{.92}
\definecolor{mycyan}{cmyk}{.3,0,0,0}
\definecolor{LightCyan}{rgb}{0.95,1,1}

\definecolor{orange}{rgb}{1,0.8,0.6}
\definecolor{yellow}{rgb}{1,1,0.6}

\newcommand{\eg}{\textit{e.g.}}
\newcommand{\ie}{\textit{i.e.}}

\usepackage[norelsize, linesnumbered, ruled, lined, boxed, commentsnumbered]{algorithm2e}
\usepackage{xcolor}
\usepackage{pifont}

\definecolor{mygray}{gray}{.9}

\usepackage{array}
\newcolumntype{x}[1]{>{\centering\arraybackslash\hspace{0pt}}p{#1}}

\makeatletter
\newcommand{\thickhline}{%
    \noalign {\ifnum 0=`}\fi \hrule height 1pt
    \futurelet \reserved@a \@xhline
}

\makeatletter
\renewcommand*{\@fnsymbol}[1]{%
  \ifcase#1\or *\or \dagger\or \ddagger\or
  \mathsection\or \mathparagraph\or \|\or **\or \dagger\dagger
  \or \ddagger\ddagger \else\@ctrerr\fi}
\makeatother

\title[Advances in Feed-Forward 3D Reconstruction and View Synthesis]%
    {Advances in Feed-Forward 3D Reconstruction \\ and View Synthesis: A Survey}

\begin{document}

\author[J. Zhang, Y. Li, et al.]
{\parbox{\textwidth}{\centering 
Jiahui Zhang$^{1}$\thanks{Equal contribution.}, \
Yuelei Li$^{2}$\footnotemark[1], \
Anpei Chen$^{3,4}$, \
Muyu Xu$^{1}$, \
Kunhao Liu$^{1}$, \
Jianyuan Wang$^{5}$, \
Xiao-Xiao Long$^{6,7}$, \\
Hanxue Liang$^{8}$, \
Zexiang Xu$^{9}$, \
Hao Su$^{4}$, \
Christian Theobalt$^{10}$, \
Christian Rupprecht$^{5}$, \
Andrea Vedaldi$^{5}$, \\
Kaichen Zhou$^{11,12}$, \
Hanspeter Pfister$^{11}$, \
Paul Pu Liang$^{12}$, \
Shijian Lu$^{1}$\thanks{Corresponding author.}, \
Fangneng Zhan$^{11,12}$
        }
\\
\\
{\parbox{\textwidth}{\centering  
$^1$NTU,
$^2$Caltech,
$^3$Westlake University,
$^4$UCSD,
$^5$University of Oxford,
$^6$Nanjing University,
$^7$HKU, \\
$^8$University of Cambridge,
$^9$Hillbot,
$^{10}$MPI for Informatics,
$^{11}$Harvard University,
$^{12}$MIT
}
}
}

\teaser{
  \vspace{-1cm}
  \includegraphics[width=0.975\linewidth]{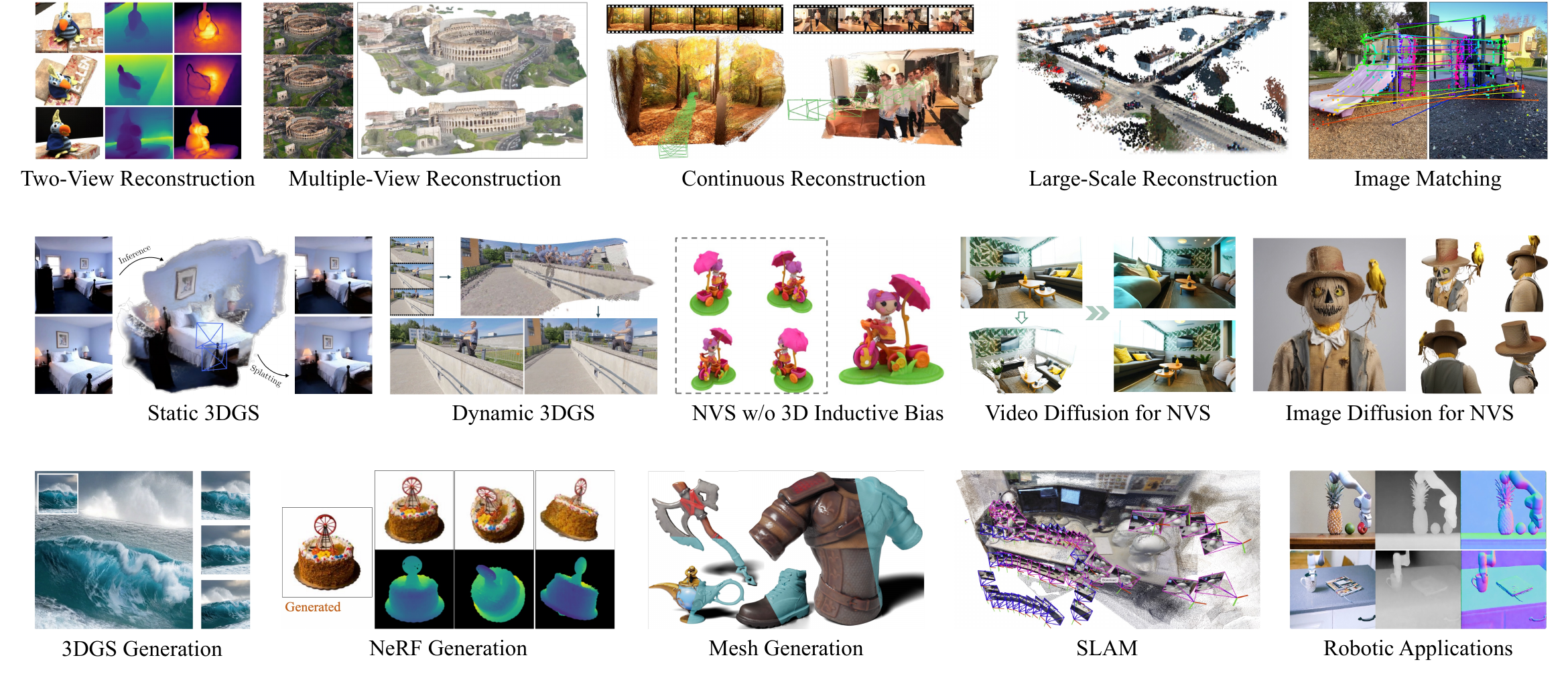}
  \centering
  \vspace{-0.3cm}
  \caption{This survey discusses various feed-forward 3D reconstruction and novel view synthesis methods.
  Images are from \cite{wang2024dust3r,wang2025vggt,wang2025continuous,leroy2024grounding,charatan2024pixelsplat,liang2024feed,jin2025lvsm,yu2024viewcrafter,szymanowicz2025bolt3d,hong2023lrm,liu2024meshformer,shi2023zero123++,deng2025vggt,maggio2025vggt,zhen2025tesseract}.
  }
  \label{fig_teaser}
}

\maketitle

\begin{abstract}
3D reconstruction and view synthesis are foundational problems in computer vision, graphics, and immersive technologies such as augmented reality (AR), virtual reality (VR), and digital twins. Traditional methods rely on computationally intensive iterative optimization in a complex chain, limiting their applicability in real-world scenarios. Recent advances in feed-forward approaches, driven by deep learning, have revolutionized this field by enabling fast and generalizable 3D reconstruction and view synthesis. 
This survey offers a comprehensive review of feed-forward techniques for 3D reconstruction and view synthesis, with a taxonomy according to the underlying representation architectures including point cloud, 3D Gaussian Splatting (3DGS), Neural Radiance Fields (NeRF), etc.
We examine key tasks such as pose-free reconstruction, dynamic 3D reconstruction, and 3D-aware image and video synthesis, highlighting their applications in digital humans, SLAM, robotics, and beyond.
In addition, we review commonly used datasets with detailed statistics, along with evaluation protocols for various downstream tasks.
We conclude by discussing open research challenges and promising directions for future work, emphasizing the potential of feed-forward approaches to advance the state of the art in 3D vision.
The project page is available at \href{https://fnzhan.com/projects/Feed-Forward-3D}{Feed-Forward-3D}.
\end{abstract}

\section{Introduction}
\label{sec:introduction}


3D reconstruction and rendering are long-standing and central challenges in computer vision and computer graphics. They enable a wide range of applications, from digital content creation, augmented reality, and virtual reality to robotics, autonomous systems, and digital twins. Traditionally, high-quality 3D reconstruction and view synthesis have relied on optimization-based pipelines such as Structure-from-Motion (SfM)~\cite{schonberger2016structure} and 3D Gaussian Splatting~\cite{kerbl20233d}.
However, these methods typically require per-scene optimization, converge slowly, and rely on dense multiview inputs, limiting their practicality in real-world scenarios. In light of this, feed-forward methods have emerged as an important research line in 3D vision.

Feed-forward models are learning-based methods that infer outputs, such as 3D geometry or novel views, in a single forward pass of a neural network, without requiring per-scene iterative optimization in classical methods. Early examples include cost-volume-based Multi-View Stereo \cite{yao2018mvsnet} and layered representations such as multiplane images (MPI) \cite{zhou2018stereo}, which demonstrate the potential of learning-based inference over per-scene optimization.
In recent years, fueled by breakthroughs in deep learning and neural representations, \emph{feed-forward methods} \cite{fan2017point,wu2016learning,yu2021pixelnerf} have emerged as a transformative alternative in 3D reconstruction and view synthesis. By learning from large scale data, these models enable orders of magnitude faster inference and better generalization compared to classical optimization methods. Their ability to produce predictions in real time makes them particularly attractive for time-sensitive and scalable applications, such as robotic perception and interactive 3D asset creation.

\begin{figure}
    \centering
\includegraphics[width=\linewidth]{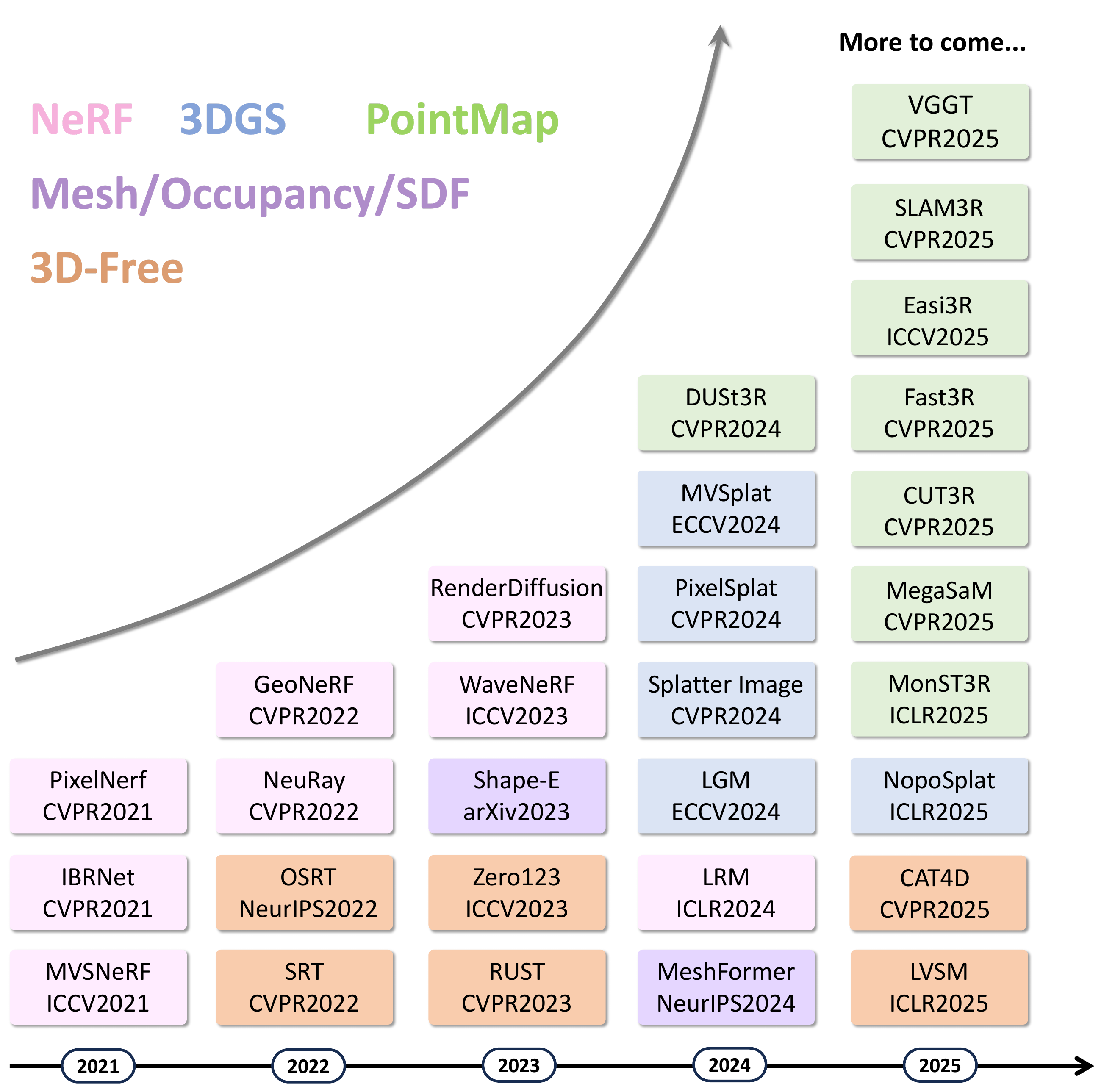}
    \caption{Timeline of representative feed-forward works categorized by methods.}
    \label{fig_summary}
\end{figure}

This survey focuses on feed-forward methods developed primarily after the emergence of neural radiance fields (NeRF)~\cite{mildenhall2021nerf} in 2020, which catalyzed a rapid evolution in feed-forward approaches, as shown in Fig.~\ref{fig_summary}.
We present a comprehensive review of feed-forward methods for 3D reconstruction and view synthesis, with an emphasis on the \textit{scene representations, core architectures, and downstream applications} that define this fast-evolving area. Specifically, we systematically categorize existing approaches based on their underlying scene representations, which determine how 3D structure and appearance are modeled and rendered. In this way, we identify five major categories: 1) models built on \textbf{Neural Radiance Fields (NeRF)}~\cite{mildenhall2021nerf}, which leverage volumetric rendering through learned radiance fields; 2) \textbf{pointmap-based} approaches~\cite{wang2024dust3r}, which operate on pixel-aligned 3D pointmaps; 3) \textbf{3D Gaussian Splatting (3DGS)}-based models~\cite{kerbl20233d}, which use rasterizable Gaussian primitives for fast and efficient rendering; 4) methods based on other 3D representations like \textbf{mesh, occupancy and signed distance function (SDF)}; and 5) \textbf{3D-free} models, which leverage deep neural networks to synthesize views directly without an explicit 3D representation.
For each category, we provide an in-depth analysis of representative and state-of-the-art methods, highlighting their core architectural designs and feature representations.

We also highlight several high-impact 3D vision applications enabled by feed-forward methods, as illustrated in Fig.~\ref{fig_teaser}. These methods offer scalable, fast, and generalizable solutions across domains, supporting tasks such as multi-view 3D reconstruction, dynamic 3D Gaussian Splatting, and diffusion-based view synthesis. Beyond reconstruction and view synthesis, they also advance image matching, 3D-aware segmentation, and optical flow estimation. In the broader context of robotics and SLAM, feed-forward models enable efficient real-time scene understanding and tracking, while in digital humans, they support generalizable avatar reconstruction from sparse inputs.

To facilitate future research, we review widely used benchmark datasets and evaluation protocols for feed-forward 3D reconstruction and view synthesis. These datasets cover synthetic and real-world scenes across objects, indoor and outdoor environments, and static or dynamic settings, with varying levels of annotation such as RGB, depth, LiDAR, and optical flow. 
We also summarize standard evaluation metrics for assessing image quality, geometry accuracy, camera pose estimation, and other relevant tasks. Together, these benchmarks and metrics provide essential foundations for comparing methods and driving progress toward more generalizable, accurate, and robust feed-forward 3D models.

Despite impressive progress, feed-forward models still face major challenges, including limited modality diversity in datasets, poor generalization in free-viewpoint synthesis, and the high computational cost of long-context processing. Addressing these challenges will require advances in efficient architectures and scalable datasets. Finally, we conclude with the societal impact of this method, highlighting the importance of responsible deployment and transparent modeling practices.

\section{Methods}
\label{methods}

We broadly categorize the feed-forward 3D reconstruction and view synthesis methods into five categories based on their underlying representation: NeRF models (Sec.~\ref{recons_nerf}),  Pointmap models (Sec.~\ref{recons_pointmap}), 3DGS models (Sec.~\ref{recons_3dgs}), models employing other common representations (\eg, mesh, occupancy, SDFs in Sec.~\ref{recons_other}), and 3D-free models (Sec.~\ref{render}).

\begin{figure*}[t]
\centering
\includegraphics[width=1\linewidth]{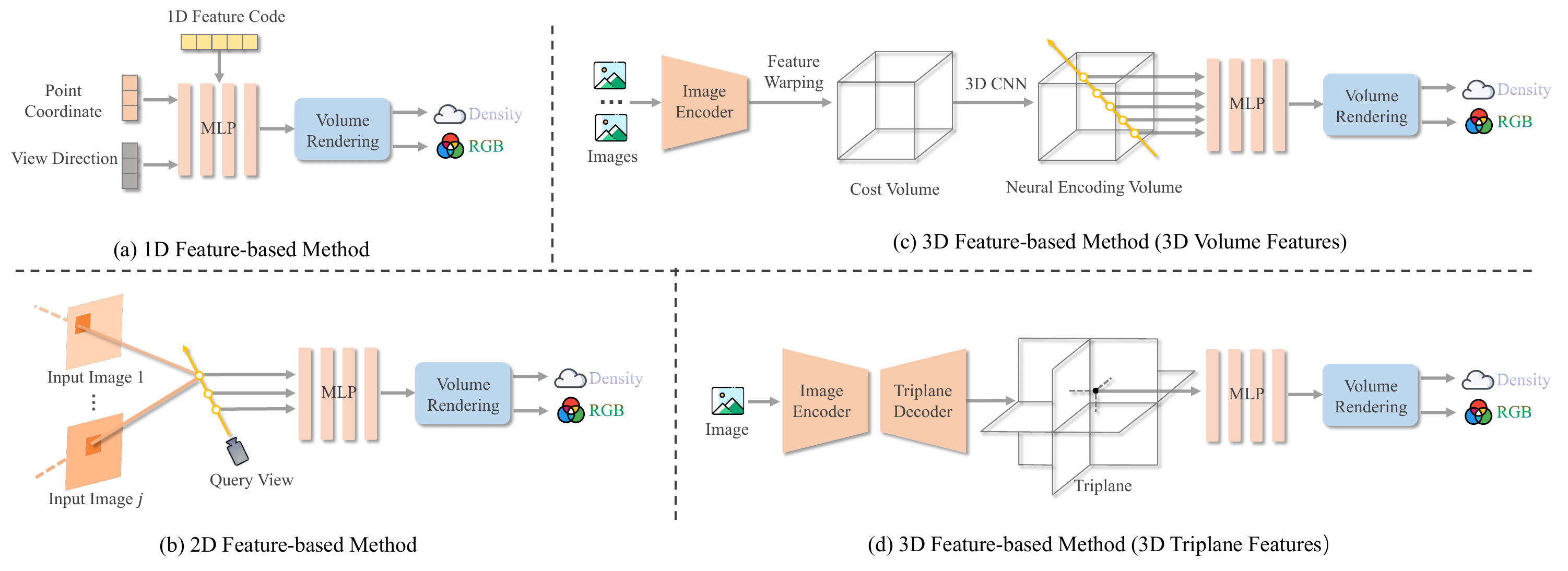}
\caption{
Representative frameworks of feed-forward NeRF, including methods based on (a) 1D features, (b) 2D features, and (c) \& (d) 3D features.
The samples are adapted from~\cite{jang2021codenerf}~\cite{trevithick2021grf}~\cite{chen2021mvsnerf}~\cite{hong2023lrm}.
}
\label{fig_recon_nerf}
\end{figure*}

\subsection{NeRF}
\label{recons_nerf}

Neural radiance fields (NeRF)~\cite{mildenhall2021nerf} have recently gained significant attention for high-quality novel view synthesis using volumetric scene representations and differentiable volume rendering. By leveraging MLPs, NeRF reconstructs 3D scenes from multiview 2D images, enabling the generation of novel views with excellent multiview consistency. However, a major limitation of NeRF is its requirement for per-scene optimization, so it cannot generalize to unseen scenes. To address this, feed-forward approaches have been proposed, where neural networks learn to infer NeRF representations directly from sparse input views, thereby eliminating the need for scene-specific optimization. As a pioneering feed-forward NeRF work, PixelNeRF~\cite{yu2021pixelnerf} introduces a conditional NeRF framework that leverages pixel-aligned image features extracted from input images, allowing the model to generalize across diverse scenes and perform novel view synthesis from sparse observations. A large number of follow-ups adopt various techniques for feed-forward NeRF, and we broadly categorize them into the following categories based on feature representations.

\subsubsection{1D Feature-based Methods }

Several methods have been proposed to encode a global 1D latent code for NeRF prediction, where the same latent code is shared between all 3D points in a scene. For example, CodeNeRF~\cite{jang2021codenerf}, as illustrated in Fig.~\ref{fig_recon_nerf}(a), introduces a disentanglement strategy that jointly learns separate embeddings for texture and shape, along with an MLP conditioned on these embeddings to predict the color and volumetric density of each 3D point. ShaRF~\cite{rematas2021sharf} introduces latent codes for shape and appearance, which serve as conditioning inputs for NeRF reconstruction.
In addition, Shap-E~\cite{jun2023shap} encodes point clouds and RGBA input images into a series of latent vectors that are subsequently used for generating NeRF's MLPs.

\subsubsection{2D Feature-based Methods } 

2D feature-based methods typically leverage an image encoder to extract image features of source views and obtain features of arbitrary 3D points by ray projection without relying on 3D intermediate features. For example, GRF~\cite{trevithick2021grf}, as illustrated in Fig.~\ref{fig_recon_nerf}(b) projects each 3D point along a camera ray onto source views to extract corresponding multiview features. These features are then aggregated and passed through an MLP to predict color and density. IBRNet~\cite{wang2021ibrnet} follows a similar approach, projecting 3D points onto nearby source views to extract image features that are aggregated across views for NeRF inference. NeRFormer~\cite{reizenstein2021common} also employs ray-projected features and performs multiview feature aggregation to guide the NeRF prediction. In addition, SRF~\cite{chibane2021stereo} projects 3D points onto multiview images to construct a stereo feature matrix, which is processed by a 2D CNN to produce view-aligned features for radiance prediction. To provide additional geometric cues for NeRF prediction, GNT~\cite{wang2022attention} introduces a view transformer that uses epipolar constraint to aggregate projected features from multiple views in a geometrically consistent manner. Its successor, GNT-MOVE~\cite{cong2023enhancing}, bridges the view transformer with the Mixture-of-Experts concept from large language models, enhancing its cross-scene generalization capability.
Instead of relying on epipolar lines, MatchNeRF~\cite{chen2023explicit} explicitly models the correspondence information by computing the similarity between ray-projected features from pairs of nearby source views, using this information as a conditioning input for NeRF prediction. In addition, ContraNeRF~\cite{yang2023contranerf} introduces geometry-aware contrastive learning~\cite{chen2020simple} to learn
multiview consistent 2D features to mitigate synthetic-to-real generalization issues.

\subsubsection{3D Feature-based Methods}
\ \quad \textbf{3D Volume Features.}~Inspired by multiview stereo (MVS)~\cite{yao2018mvsnet, gu2020cascade, cheng2020deep}, MVSNeRF~\cite{chen2021mvsnerf} constructs 3D cost volumes that store the matching costs of pixels/features across multiple images from input images (usually along depth values).
These cost volumes are used to generate a neural scene encoding volume, which stores per-voxel features representing both local geometry and appearance, as shown in Fig.~\ref{fig_recon_nerf}(c). For any 3D point, its features are obtained via trilinear interpolation from the encoding volume and then decoded by an MLP to predict the corresponding density and color. To improve rendering quality in both fine-detail areas and occluded regions, GeoNeRF~\cite{johari2022geonerf} extends MVSNeRF by first constructing cascaded 3D cost volumes for each source view, followed by an attention-based volume aggregation across views. 
Similarly addressing occlusions, NeuRay~\cite{liu2022neural} leverages cost volumes to predict the visibility of 3D points, which can identify feature inconsistencies caused by occlusion. 
WaveNeRF~\cite{xu2023wavenerf} further enhances geometry reconstruction by integrating wavelet frequency volumes into the MVS pipeline to preserve high-frequency details.

For efficient rendering, 
ENeRF~\cite{lin2022efficient} samples a limited number of points near the scene surface, guided by coarse scene geometry predicted from a cascaded cost volume, thereby speeding up rendering;
Instead of constructing cost volumes for all reference input views, MuRF~\cite{xu2024murf} efficiently constructs a target view frustum volume to aggregate information from the input images;
To improve the quality of geometry estimation, GeFu~\cite{liu2024geometry} introduces an adaptive cost aggregation module that learns to reweight contributions from different source views, enabling more accurate cost volume construction.

\textbf{3D Triplane Features.} 
The 3D triplane serves as an efficient volumetric representation \cite{peng2020convolutional,chan2022efficient}, making it highly compatible with feed-forward models.
Specifically, Large Reconstruction Model (LRM)~\cite{hong2023lrm} employs a large transformer-based encoder-decoder architecture and directly regresses a feature triplane representation as shown in Fig.~\ref{fig_recon_nerf}(d), enabling NeRF prediction from triplane features. Pf-LRM~\cite{wang2023pf} extends LRM to a pose-free setting, which jointly reconstructs the triplane NeRF representations and predicts relative camera poses. TripoSR~\cite{tochilkin2024triposr} further advances LRM by carefully curating a subset of Objaverse~\cite{deitke2023objaverse} for training, along with improved architectures and training strategies. 
To address the scarcity, licensing constraints, and inherent biases of 3D data, LRM-Zero~\cite{xie2024lrm} introduces the Zeroverse dataset and performs training entirely on synthesized data. Beyond reconstruction alone, several methods integrate LRM with diffusion models. For example, Instant3D~\cite{li2023instant3d} first leverages a fine-tuned 2D diffusion model~\cite{podell2023sdxl} to generate 4-view images from a text prompt and then uses a transformer-based large reconstruction model to predict a NeRF; DMV3D~\cite{xu2023dmv3d} incorporates LRM into multiview diffusion, which gradually reconstructs a clean triplane NeRF representation from noisy multiview images in the diffusion process.

\subsubsection{Other Methods.} In addition to the methods discussed above, several approaches have explored feed-forward NeRF reconstruction with other types of features. For example, VisionNeRF~\cite{lin2023vision} proposes to leverage vision transformer~\cite{dosovitskiy2020image} and convolutional networks to extract global 1D features and 2D image features, respectively, and constructs a multi-level feature map that serves as the conditioning inputs of NeRF prediction to enhance rendering quality, particularly in occluded regions. MINE~\cite{li2021mine} integrates NeRF and multiplane image (MPI)~\cite{tucker2020single} representations to enable generalizable, occlusion-aware 3D reconstruction from a single image.

\subsection{Pointmap}
\label{recons_pointmap}

\begin{figure*}[t]
\centering
\includegraphics[width=0.95\linewidth]{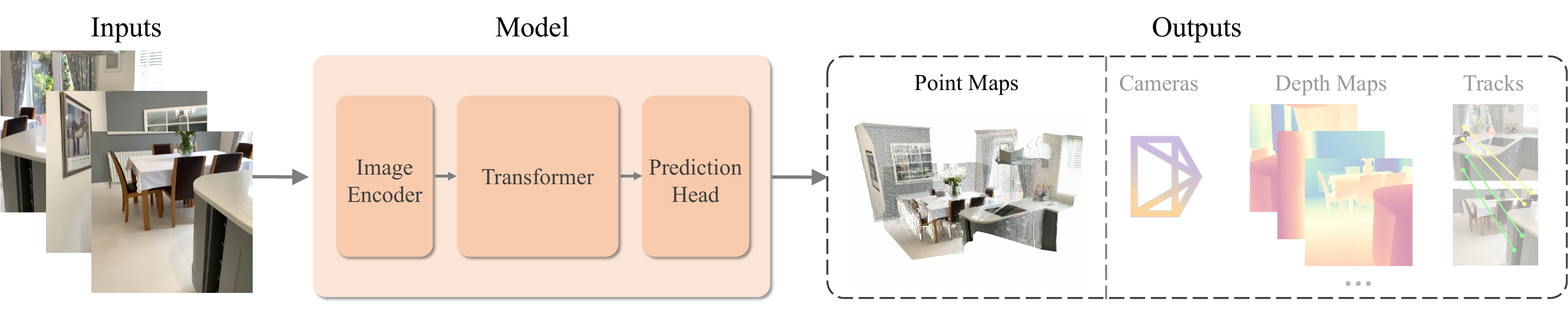}
\caption{
The framework of feed-forward pointmap reconstruction. It also supports broader tasks, such as camera estimation, depth estimation, and point tracking, within the same framework.
The samples are adapted from \cite{wang2025continuous}.
}
\label{fig_recon_pointmap}
\end{figure*}

Pointmaps~\cite{wang2024dust3r, brachmann2017dsac, brachmann2018learning, brachmann2021visual, dong2022visual}, 
encode scene geometry, pixel-to-scene correspondences, and viewpoint relationships, allowing for camera poses, depths, and explicit 3D primitive estimation as shown in Fig. \ref{fig_recon_pointmap}. The pioneering feed-forward pointmap reconstruction method DUSt3R~\cite{wang2024dust3r} learns a transformer-based encoder-decoder to directly output two pixel-aligned pointmaps from image pairs without posed cameras, enabling dense unconstrained stereo 3D reconstruction. The follow-up work, MASt3R~\cite{leroy2024grounding}, improves DUSt3R by introducing local feature matching.

To handle more views, Fast3R~\cite{yang2025fast3r} builds on DUSt3R and designs a global fusion transformer to process multiview inputs simultaneously.
MV-DUSt3R~\cite{tang2025mv} instead leverages multiview decoder blocks to learn both the reference-to-source and source-to-source view relationships, thereby extending DUSt3R to a multiview setting.
SLAM3R~\cite{liu2025slam3r} introduces an Image-to-Points module that enables simultaneous processing of multiview inputs, effectively enhancing reconstruction quality without sequential reconstruction.

%
Notably, several works incorporate a memory mechanism that incrementally processes inputs and updates a scene’s latent state, progressively adding points to a canonical 3D representation.
Spann3R~\cite{wang20243d} introduces a spatial memory network that enables multiview inputs while improving efficiency, removing the need for global alignment. 
Similarly,
MUSt3R~\cite{cabon2025must3r} extends the DUSt3R architecture with a symmetric design and a memory mechanism, effectively reducing computational complexity when processing multiview inputs. 
CUT3R~\cite{wang2025continuous} proposes a Continuous Updating Transformer that simultaneously updates the state with new information and retrieves the information stored in the state. This formulation is general, supporting video and photo collections as well as static and dynamic scenes.
However, with the increased number of processed frames, memory-based methods face capacity constraints, which can result in the degradation or loss of information from earlier frames. To address this issue, Point3R~\cite{wu2025point3r} takes inspiration from the human memory mechanism and proposes a spatial pointer memory, where each pointer is anchored at a 3D position and links to a dynamically evolving spatial feature. 
Driv3R~\cite{fei2024driv3r} extends the memory mechanism to support efficient temporal integration, enabling large-scale dynamic scene reconstruction from multiview input sequences.

In addition, several methods are proposed to develop new SfM pipelines for efficient 3D reconstruction. Specifically, Light3R-SfM~\cite{elflein2025light3r} replaces optimization-based global alignment with a learnable latent alignment module, enabling the efficient SfM and 3D reconstruction. Regist3R~\cite{liu2025regist3r} introduces a stereo foundation model to build a scalable incremental SfM pipeline for efficient 3D reconstruction.

To facilitate accurate 3D reconstruction, Pow3R~\cite{jang2025pow3r} flexibly integrates available priors at test time, such as camera intrinsics, sparse or dense depth, or relative poses, as lightweight and diverse conditioning. In contrast, Rig3R~\cite{li2025rig3r} exploits the rig metadata as conditions to improve both the camera pose estimation and 3D reconstruction. Meanwhile, MoGe~\cite{wang2025moge} improves geometry learning by replacing DUSt3R’s scale-invariant pointmaps with affine-invariant ones, and further introduces a global alignment solver to correct scale and shift errors in affine-invariant pointmaps. Similarly focused on geometric consistency, Test3R~\cite{yuan2025test3r} takes advantage of test time training to improve the geometric consistency of pointmaps. AerialMegaDepth~\cite{vuong2025aerialmegadepth} instead focuses on aerial-ground geometric reconstruction from a data perspective.

As a promising and powerful foundation for 3D reconstruction, VGGT~\cite{wang2025vggt} presents a large feed-forward transformer-based architecture that directly predicts all essential 3D attributes, such as camera intrinsics and extrinsics, point maps, depth maps, and 3D point tracks, without the need for post-processing, leading to state-of-the-art 3D point and camera pose reconstruction. VGGT-Long~\cite{deng2025vggt} extends VGGT to handle kilometer-scale sequences through a chunk-and-align pipeline, effectively mitigating its memory constraints in long-sequence handling.

\subsection{3DGS}
\label{recons_3dgs}

3D Gaussian Splatting (GS)~\cite{kerbl20233d} is a recent advance for efficient 3D reconstruction and rendering built on rasterization. 
%
Despite its high fidelity in reconstruction, 3DGS requires per-scene optimization, which limits its training efficiency and generalization capabilities. 
Recently, feed-forward 3DGS reconstruction methods have been developed by leveraging neural networks to directly predict Gaussian parameters, eliminating the need for per-scene optimization.
%
We categorize these methods based on the representation of predicted Gaussian outputs: 2D map and 3D volume.

\begin{figure*}[t]
\centering
\includegraphics[width=1\linewidth]{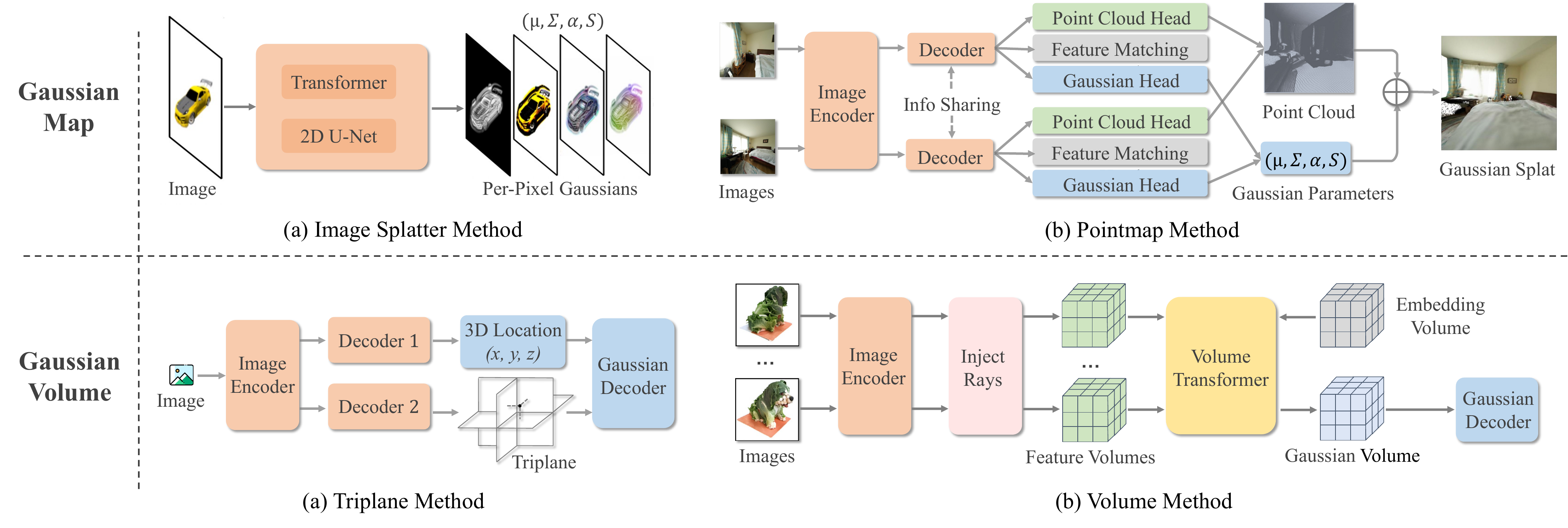}
\caption{
Representative frameworks with different outputs of 3D Gaussian representations, including Gaussian Map (e.g., Image Splatter and pointmap) and Gaussian Volume (e.g., Triplane and 3D Volume).
The samples are adapted from~\cite{szymanowicz2024splatter}~\cite{chen2024lara}~\cite{smart2024splatt3r}~\cite{zou2024triplane}.
}
\label{fig_recon_3dgs}
\end{figure*}

\subsubsection{Gaussian Map}
A Gaussian map refers to a 2D-based representation of 3D Gaussians, where each 2D location encodes a 3D Gaussian.
As a pioneering effort, Splatter Image~\cite{szymanowicz2024splatter} employs a U-Net encoder-decoder architecture~\cite{ronneberger2015u} to predict pixel-aligned 3D Gaussians for single-view 3D object reconstruction as illustrated in Fig.~\ref{fig_recon_3dgs}.
To improve the reconstruction quality, subsequent methods leverage generic scene priors learned from large-scale datasets.  
GRM~\cite{xu2024grm} directly maps input image pixels to pixel-aligned 3D Gaussians for feed-forward object reconstruction, based on LRM~\cite{hong2023lrm} that learns general reconstruction priors from large-scale datasets of 3D objects;
Flash3D~\cite{szymanowicz2024flash3d} further introduces a high-quality depth predictor as prior to achieve single-view scene-level reconstruction.
Concurrently, GS-LRM~\cite{zhang2024gs} incorporates Transformer-based LRM to formulate per-pixel Gaussian prediction as a sequence-to-sequence mapping, achieving remarkable performance across both objects and scenes. 
This line of research has been further advanced by several works: eFreeSplat~\cite{min2024epipolar} leverages a large vision transformer encoder~\cite{dosovitskiy2020image} as 3D priors for Gaussian image prediction; Long-LRM~\cite{ziwen2024long} that combines Mamba2 blocks~\cite{dao2024transformers} with Transformer layers to handle long sequences of input images; and FreeSplatter~\cite{xu2024freesplatter} achieves pose-free Gaussian map prediction.
However, these works are limited to reconstructing existing image observations without generative capabilities.
To overcome this limitation, LGM~\cite{tang2024lgm} introduces pre-trained diffusion models~\cite{long2024wonder3d, shi2023zero123++, shi2023mvdream, wang2023imagedream} to generate multiview images, which are then used by large multiview Gaussian models for multiview Gaussian prediction.
Wonderland~\cite{liang2025wonderland} further enhances this approach by leveraging a pre-trained video diffusion model~\cite{yang2024cogvideox} to generate informative video latents from a single image for 3DGS prediction.
In addition to the methods above, another line of work focuses on the geometric quality of 3D scene reconstruction by incorporating geometric designs or priors, such as epipolar constraints, cost volumes, and pre-trained reconstruction models.

\noindent\textbf{Epipolar-based Methods.} As a pioneering epipolar-based method, PixelSplat~\cite{charatan2024pixelsplat} leverages an epipolar line to resolve the scale ambiguity issue and capture cross-view features, estimating a probabilistic depth distribution as 3D Gaussian positions. Although PixelSplat is effective in regions strongly correlated with the input observations, it struggles in areas of high uncertainty, leading to blurry or failed reconstructions.
To address this, LatentSplat~\cite{wewer2024latentsplat} combines the strengths of regression-based and generative approaches to obtain high-quality reconstructions in uncertain areas. 
Building further on PixelSplat, GGRt~\cite{li2024ggrt} introduces a joint learning framework that integrates pose estimation with a 3DGS prediction network to enable pose-free 3D Gaussian prediction.

\noindent\textbf{Cost Volume-based Methods.} A key limitation of PixelSplat is the inherent ambiguity and unreliability in mapping image features to depth distributions, resulting in suboptimal geometry reconstruction. To address this problem, MVSplat~\cite{chen2024mvsplat} and MVSGaussian~\cite{pham2024mvgaussian} adopt a plane-sweeping-based cost volume to facilitate multiview Gaussian image prediction, leveraging cross-view feature similarities to improve depth estimation. 
However, these methods heavily depend on precise multiview feature matching, which becomes particularly challenging in scenes with occlusions, low texture, or repetitive patterns. To address this issue, TranSplat~\cite{zhang2025transplat} introduces a depth-aware deformable matching transformer to generate a depth confidence map to facilitate multiview feature matching.
Similarly, DepthSplat~\cite{xu2024depthsplat} uses a multiview depth model that leverages pre-trained monocular depth features to enhance the feed-forward 3DGS reconstruction. 
This line of research is further extended by several works. HiSplat~\cite{tang2024hisplat} introduces hierarchical representations to capture large-scale structures and fine texture details; PanSplat~\cite{zhang2024pansplat} builds a hierarchical spherical cost volume for 4K panorama view synthesis; MVSplat360~\cite{chen2024mvsplat360} adapts MVSplat to support 360° novel view synthesis for large-scale scenes; and LongSplat~\cite{huang2025longsplat} designs an online updating framework for long-sequence image streams.

\noindent\textbf{Pre-trained Reconstruction Models.} 
With the advent of feed-forward pointmap reconstruction methods~\cite{wang2024dust3r, leroy2024grounding, wang20243d} as mentioned in Sec.~\ref{recons_pointmap}, several feed-forward 3DGS methods directly use pre-trained models to generate dense pointmaps for 3D Gaussian reconstruction. For example, Splatt3R~\cite{smart2024splatt3r} builds on the large-scale pretrained foundation 3D MASt3R model~\cite{leroy2024grounding} by integrating a Gaussian decoder, enabling pose-free feed-forward 3DGS. NoPoSplat~\cite{ye2024no} also uses MASt3R as the backbone and predicts 3D Gaussians in a canonical space without ground-truth camera poses and depth. 
SmileSplat~\cite{li2024smilesplat} instead uses DUSt3R as the backbone to predict Gaussian surfels with a multi-head Gaussian regression decoder. SelfSplat~\cite{kang2024selfsplat} unifies Gaussian prediction with self-supervised learning of depth and camera poses, enabling simultaneous prediction of geometry, pose, and Gaussian attributes. 
However, methods relying on DUSt3R and MASt3R inherit the limitation of pairwise inputs, which restricts their scalability.
Thus, PREF3R~\cite{chen2024pref3r} adopts Spann3R~\cite{wang20243d} as the geometric prior, which introduces a spatial memory network to handle multiview images.
%
Except for feed-forward pointmap prediction models, pre-trained 3D diffusion models~\cite{jun2023shap} have also been explored as a geometric prior for 3D Gaussian prediction~\cite{lu2024large}. 

\subsubsection{Gaussian Volume}

Gaussian volume~\cite{chen2024lara} represents 3D with Gaussian voxel grids, where each voxel comprises multiple Gaussian primitives. A typical feed-forward 3DGS method using a Gaussian volume representation is LaRa~\cite{chen2024lara}, which aims to reduce the heavy training cost associated with 360° bounded radiance field reconstruction. As shown in Fig.~\ref{fig_recon_3dgs} (b), it first builds 3D features and embedding volumes and then leverages a volume transformer to reconstruct a Gaussian volume, enabling progressively and implicitly feature matching that leads to higher quality results and faster convergence. Building on the notion of structured volumetric Gaussians, GaussianCube~\cite{zhang2024gaussiancube} proposes a structured and explicit radiance representation for 3D object generation from a single image.
In parallel, QuickSplat~\cite{liu2025quicksplat}, based on G3R~\cite{chen2024g3r}, predicts voxel-level features using neural networks, which are subsequently decoded into Gaussian primitives with MLPs for surface reconstruction.
To further enrich volumetric representations, GD~\cite{nam2025generative} builds upon LaRa and introduces a generative densification that exploits prior knowledge from large multiview datasets and densifies feature representations from feed-forward 3DGS. SCube~\cite{ren2024scube} advances this direction toward large-scale scene reconstruction by proposing VoxSplat, a high-resolution sparse-voxel Gaussian representation generated via a hierarchical latent diffusion model conditioned on sparse posed images. 
Most recently, AnySplat~\cite{jiang2025anysplat} constructs volumetric Gaussians by voxelizing the Gaussian maps from variable numbers of input views, yielding a unified and efficient model capable of handling both sparse and dense views.

To mitigate the high computational cost of the 3D volume, an efficient triplane structure~\cite{peng2020convolutional,chan2022efficient} is also explored for Gaussian volume prediction.
It is typically constructed by predicting a triplane representation first and then leveraging the latent triplane features to decode 3D Gaussians, as illustrated in Fig.~\ref{fig_recon_3dgs}. For example, Triplane-Gaussian~\cite{zou2024triplane} leverages several transformer-based networks pre-trained in large-scale datasets to build a Gaussian triplane, enabling high-quality single-view 3D reconstruction. AGG~\cite{xu2024agg} also mixes triplane and 3D Gaussians, which first represents scene textures as triplane and then decodes 3D Gaussians from triplane-based texture features queried by 3D locations.

\subsection{Other 3D Representations}
\label{recons_other}

Except for the methods mentioned above, there have been several efforts dedicated to the feed-forward reconstruction with different 3D representations. In this section, we introduce several representative types, including methods based on the mesh, occupancy, and signed distance function (SDF).

\subsubsection{Mesh}
Meshes are compatible with various graphics pipelines and have gained significant attention in feed-forward 3D reconstruction in recent years. For instance, Pixel2Mesh~\cite{wang2018pixel2mesh} generates a 3D mesh from a single image by extracting features with a 2D CNN and progressively deforming an initial mesh; Mesh R-CNN~\cite{gkioxari2019mesh} introduces a voxel branch based on Mask R-CNN~\cite{he2017mask} to predict a coarse cubified mesh followed by further refinement. 

This field has been further advanced by the development of diffusion models. For example, One-2-3-45~\cite{liu2023one} builds on the diffusion-based model to generate multiview images, which are then processed by a generalizable surface reconstruction module~\cite{long2022sparseneus} for mesh reconstruction;
One-2-3-45++~\cite{liu2024one} further utilizes a 3D diffusion-based module conditioned on multiple views to generate a textured mesh in a coarse-to-fine manner. 
To improve geometry consistency, Wonder3D~\cite{long2024wonder3d} introduces a cross-domain diffusion model to generate multiview-consistent normal maps and RGB images simultaneously, which can be used for mesh reconstruction. Similarly, Unique3D~\cite{wu2024unique3d} employs a multiview diffusion model alongside a normal diffusion model to generate multiview-consistent images and normal maps, which are fed into a fast and consistent mesh reconstruction module.

In addition, several methods are proposed to take advantage of large reconstruction model~\cite{hong2023lrm} to achieve high-quality mesh reconstruction. For example, MeshLRM~\cite{wei2024meshlrm} integrates differentiable surface extraction and rendering into a large reconstruction model, enabling direct 3D mesh generation;
InstantMesh~\cite{xu2024instantmesh} utilizes a transformer-based large reconstruction model to generate a high-quality 3D mesh from multiview images;
MeshFormer~\cite{liu2024meshformer} leverages 3D voxel representations and combines transformers with 3D convolutions to improve 3D mesh geometry.

To generate artist-created meshes with high-quality topology, several methods draw inspiration from large language models, treating 3D meshes as sequences and introducing autoregressive transformer architectures tailored to this sequential representation. Specifically, MeshGPT~\cite{siddiqui2024meshgpt} leverages VQVAE~\cite{van2017neural} to learn a mesh vocabulary and employs a decoder-only transformer to autoregressively generate triangle meshes. To mitigate cumulative errors inherent in VQVAE-based methods, MeshXL~\cite{chen2024meshxl} further introduces a neural coordinate field for sequential 3D mesh representation.
To address the problem that the above methods struggle to learn the shape and topology distributions together, MeshAnything~\cite{chen2024meshanything} uses a pre-trained encoder~\cite{zhao2023michelangelo} to inject shape features into a VQVAE-based sequence, removing the need to learn shape distribution and enabling focused topology learning.

\subsubsection{Occupancy} Occupancy~\cite{mescheder2019occupancy, peng2020convolutional} refers to the property that describes whether a given point in a 3D space is inside or outside a surface or object. Several methods have been proposed to achieve feed-forward occupancy prediction from images. For example, 
Any-Shot GIN~\cite{xian2022any} aims to model occupancy-based 3D implicit reconstruction. It begins with front-back depth estimation to generate depth maps for constructing a voxel-based representation, subsequently extracting 3D features from this volume to infer the occupancy of any 3D point in the space;
MCC~\cite{wu2023multiview} encodes a compressed representation of the scene appearance and geometry, and then uses the representation to predict occupancy probability and RGB color for each 3D point;
ZeroShape~\cite{huang2024zeroshape} employs intermediate geometric representation and explicit reasoning to achieve 3D occupancy regression.
While there are also many works on semantic occupancy prediction for autonomous driving, we do not cover these methods here, as the focus of this survey is on 3D reconstruction and view synthesis.

\subsubsection{SDF} Signed Distance Function (SDF)~\cite{park2019deepsdf} is a mathematical function that represents the geometry of a shape or surface in space. For any point in 3D space (or 2D), the SDF returns the shortest distance from that point to the surface of the object. The sign of the distance indicates whether the point is inside or outside the object. Several methods have been proposed to enable feed-forward SDF representations. 
For example, 
SparseNeuS~\cite{long2022sparseneus} initially builds a hierarchy of volumes that represent local surface details, which are then used to infer SDF-based surfaces through a progressive coarse-to-fine process;
Shap-E~\cite{jun2023shap} transforms point clouds and RGBA input images into a sequence of latent vectors for SDF prediction. 
Some works also incorporate multiview stereo (MVS) for SDF prediction. For instance, C2F2NeUS~\cite{xu2023c2f2neus} constructs a hierarchy of geometric frustums to capture local-to-global geometry for SDF prediction; UFORecon~\cite{na2024uforecon} introduces a cross-view matching transformer to extract cross-view matching features to construct hierarchical correlation volumes, enabling SDF-based reconstruction with limited camera view overlaps.
Without relying on MVS, CRM~\cite{wang2024crm} predicts SDF by incorporating geometric priors into network designs based on the spatial alignment between triplanes and six input orthographic views generated by a multiview diffusion model.
There have also been works that explore transformers for SDF prediction,
including VolRecon~\cite{ren2023volrecon} which employs 
a view transformer to integrate cross-view features and a ray transformer to estimate SDF values, and ReTR~\cite{liang2023retr} which introduces an occlusion transformer and a render transformer to fuse features and perform rendering.

\begin{figure}[t]
\centering
\includegraphics[width=1\linewidth]{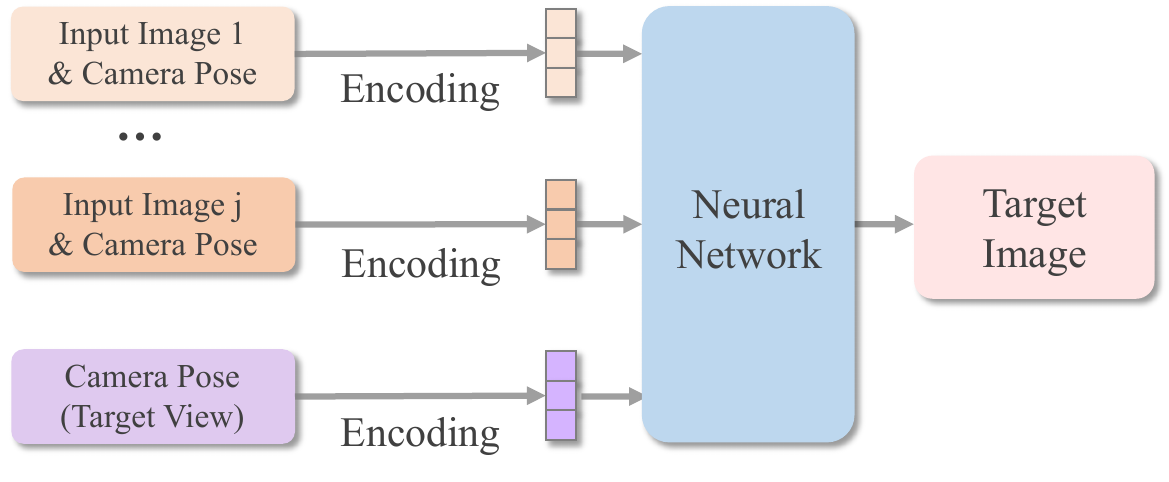}
\caption{
Typical frameworks of regression-based 3D-free models. 
The image is adapted from ~\cite{jin2025lvsm}.
}
\label{fig_render_regression}
\end{figure}

\subsection{3D-Free Models}
\label{render}

Feed-forward 3D-free models aim to directly synthesize novel views without 3D representations (e.g., NeRF and 3DGS). We broadly categorize the methods into two categories: regression-based methods (Sec.~\ref{render_transformer}) and generative methods (Sec.~\ref{render_diffusion}).

\subsubsection{Regression-based View Synthesis}
\label{render_transformer}

Regression-based methods aim to formulate the rendering process as a regression problem, learning a rendering function (typically a transformer-based network) to predict the pixel colors of novel views from sparse-view inputs directly, eliminating the inductive bias inherent in 3D representations as shown in Fig.~\ref{fig_render_regression}.

Scene representation transformer (SRT)~\cite{sajjadi2022scene} leverages a transformer-based encoder to map multiview input images to latent representations first and then outputs novel-view images from a transformer-based decoder with light field rays. RUST~\cite{sajjadi2023rust} inherits an encoder-decoder architecture and enables novel view synthesis solely from RGB images, without the need for camera poses. 
With a focus on object-centric 3D scenes, OSRT~\cite{sajjadi2022object} incorporates a slot attention module on SRT to map the encoded latent representations to object-centric slot representations.
To extend SRT to large-scale scenes, RePAST~\cite{safin2023repast} integrates relative camera pose information into the attention layer of SRT. However, these methods often suffer from degraded details and suboptimal rendering quality. 
To address this issue, several approaches incorporate geometric information to improve model performance. 
For example, GPNR~\cite{suhail2022generalizable} integrates epipolar geometry within its encoder-decoder architecture, while Du et al.~\cite{du2023learning} introduce a multiview vision transformer and epipolar line sampling to improve scene geometry. GBT~\cite{venkat2023geometry} incorporates ray distance-based geometry reasoning into multihead attention layers of transformers in the encoder and decoder.
GTA~\cite{miyato2023gta} introduces geometric transform attention to embed the geometrical structure of tokens into the transformer and integrates it into SRT to enhance transformer-based rendering. 
However, despite the improved model performance, geometrical designs often integrate additional 3D inductive biases.
LVSM~\cite{jin2025lvsm} removes the geometrical designs and leverages a transformer-based large reconstruction model with self-attention to regress the target view pixels.
To enable novel view synthesis without ground truth of cameras, RayZer~\cite{jiang2025rayzer} introduces a self-supervised multiview model that first learns camera parameters and latent scene representations from unposed input images, and then renders novel views.

\begin{figure*}[ht]
\centering
\includegraphics[width=0.95\linewidth]{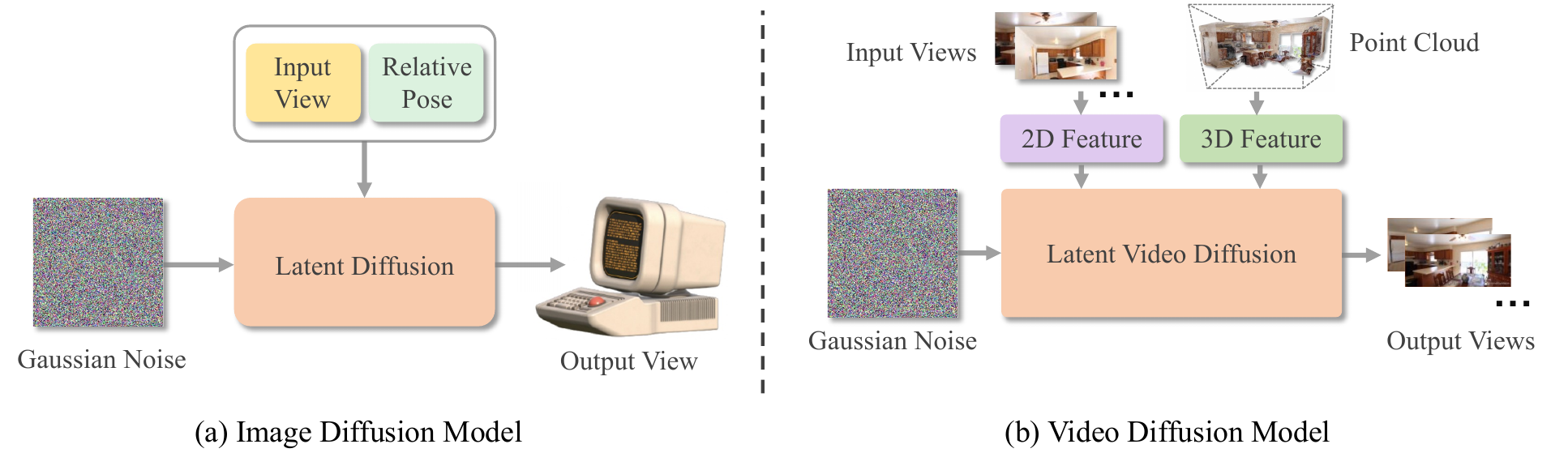}
\caption{
Representative frameworks of generative 3D-free models. The samples are adapted from ~\cite{liu2023zero} and ~\cite{liu2024reconx}.
}
\label{fig_render_generative}
\end{figure*}

\subsubsection{Generative View Synthesis}
\label{render_diffusion}

Regression-based methods work well for view interpolation, while it struggles with view extrapolation, e.g., estimating unseen regions of the scene.
In contrast, generative methods instead leverage generative models to synthesize realistic novel views based on learned data distributions, enabling view extrapolation even from a single input image.

Earlier works primarily used transformer-based autoregressive models~\cite{van2017neural, esser2021taming}. For example, GFVS~\cite{rombach2021geometry} approaches novel view synthesis by sampling target images from a learned distribution conditioned on a source image and camera transformation, where the distribution is modeled autoregressively with a transformer.
ViewFormer~\cite{kulhanek2022viewformer} further extends single-view NVS of GFVS to multiview NVS through a proposed branching attention.
Recently, latent diffusion models~\cite{rombach2022high} have been widely used in novel view synthesis due to their ability in generating high-resolution images. 
These models encode inputs into a latent space using a pretrained variational autoencoder and perform diffusion within the latent space. We categorize them into latent image diffusion models and video diffusion models.

\textbf{Image Diffusion Models.} Zero-1-to-3~\cite{liu2023zero} leverages the latent diffusion model~\cite{rombach2022high} pretrained for text-to-image generation and replaces text embedding with relative camera poses as conditioning to achieve novel view synthesis as illustrated in Fig.~\ref{fig_render_generative}(a). ZeroNVS~\cite{sargent2024zeronvs} extends Zero-1-to-3 to achieve single-view scene-level novel view synthesis, with an additional field of view as camera parameters to solve the scale ambiguity.
However, these methods still face challenges in generating consistent novel views. SyncDreamer~\cite{liu2023syncdreamer} addresses this by initializing a diffusion model with Zero-1-to-3 weights and modeling the joint distribution of multiview images. Zero123++\cite{shi2023zero123++} arranges six views in a single image for improved multiview representation modeling. ViewDiff~\cite{hollein2024viewdiff} also exploits diffusion priors, with an emphasis on multiview consistency in real-world data. 
Consistent123~\cite{weng2023consistent123} combines Zero-1-to-3 and stable diffusion to provide case-aware diffusion priors to ensure multiview-consistency, while ConsistNet~\cite{yang2024consistnet} performs parallel viewpoint-specific diffusion and aligns the generated images accordingly to enforce multiview geometric consistency. MVDream~\cite{shi2023mvdream} proposes a multiview diffusion model that leverages both 2D and 3D data, combining the generalizability of 2D diffusion models with the consistency of 3D renderings. 
With similar multiview diffusion architectures, CAT3D~\cite{gao2024cat3d} \& SEVA~\cite{zhou2025stable} and CAT4D~\cite{wu2025cat4d} substantially advance the performance for static and dynamic view synthesis.

\textbf{Video Diffusion Models.} Video diffusion models~\cite{ho2022video, xing2024dynamicrafter} have achieved impressive performance in video synthesis and are believed to implicitly capture 3D structures. Building on this capability, recent approaches have explored leveraging video diffusion priors to generate multiview images for high-quality 3D reconstruction. For example, ReconX~\cite{liu2024reconx}, shown in Fig.~\ref{fig_render_generative}(b), uses the generative prior of large pretrained video diffusion models~\cite{xing2024dynamicrafter} to synthesize novel views. It encodes extracted point clouds as 3D structural conditions, ensuring multiview consistency in the generated novel views. 
Instead of directly using point clouds as condition, ViewCrafter~\cite{yu2024viewcrafter} renders point cloud into images as conditions of the video diffusion model~\cite{xing2024dynamicrafter} to enable consistent and accurate novel view synthesis. MultiDiff~\cite{muller2024multidiff} leverages a single reference image and a predefined target camera trajectory as conditions of diffusion models, with depth cues to encourage consistent novel view synthesis. More recently, DifFusion3D+~\cite{wu2025difix3d}, SpatialCrafter~\cite{zhang2025spatialcrafter}, and GenFusion~\cite{Wu2025GenFusion} have bridged reconstruction and generation by using video diffusion models as scene reconstruction refiners, enabling both artifact removal and scene content expansion.

\section{Tasks \& Applications}
The preceding methods section highlights the core architectures of feed-forward models, 
which also include the corresponding basic tasks of novel view synthesis or static 3D reconstruction.
In contrast, this section extends beyond the basic tasks and explores the wide variety of applications enabled by feed-forward models.
Notably, we do not cover tasks and applications centered on depth, as this is an independent research line, even though depth is sometimes an output of feed-forward models.

\subsection{Camera Pose Estimation}
Feed-forward models that predict pointmaps enable efficient recovery of camera parameters without the need for explicit multi-stage SfM pipelines. DUSt3R demonstrates how pose estimation tasks can be naturally derived from pointmap output.
Specifically, camera intrinsics such as the focal length can be solved by minimizing reprojection errors of pointmaps.
The relative camera motion between two views can be recovered either by first performing pixel matching and intrinsics estimation to compute the essential matrix~\cite{hartley2003multiple}, or more directly by aligning the predicted pointmaps across views through Procrustes alignment~\cite{luo1999procrustes}, which yields rotation and translation up to scale. To improve robustness against noise and outliers, pointmap alignment can also be integrated with PnP-RANSAC~\cite{fischler1981random,hartley2003multiple,lepetit2009ep}.

Instead of relying on pointmaps to derive camera parameters, another line of research \cite{zhang2025flare,wang2025continuous,wang2025vggt} aims to predict camera parameters directly in a feed-forward manner. Typically, a learnable camera token is prepended to the image tokens, which allows for interaction with other views to capture image-level information, e.g., camera motions.
On the other hand, most of them anchor their pose estimation to a reference viewpoint like the first frame, which is an inductive bias that can lead to inferior results when the reference is suboptimal. 
$\pi^{3}$~\cite{wang2025pi} instead proposes to estimate the relative camera pose between views, achieving superior performance of camera pose estimation and benefiting other associated tasks such as pointmap estimation.


\subsection{Pose-Free 3D Reconstruction \& View Synthesis} 
The development of feed-forward models has enabled the reconstruction of 3D scenes from unposed images or videos without the need for per-scene optimization. FlowCam~\cite{smith2023flowcam} uses a single-view feed-forward NeRF combined with optical flow to estimate poses and fuse multiview point maps for NeRF-based 3D reconstruction. CoPoNeRF~\cite{hong2024unifying} builds and refines 4D correlation maps from image pairs to estimate flow and poses, enabling color and depth rendering.

To extend these approaches to 3DGS, GGRt~\cite{li2024ggrt} employs PixelSplat~\cite{charatan2024pixelsplat} for predicting viewpoint-specific 3D Gaussian maps and introduces a pose estimation module that jointly optimizes camera poses alongside Gaussian predictions. PF3plat~\cite{hong2024pf3plat} proposes a coarse-to-fine strategy, estimating depth, confidence, and camera poses from input images to guide the prediction of 3D Gaussians.

Additionally, several methods build upon DUSt3R~\cite{wang2024dust3r} for pose-free 3D reconstruction. DUSt3R itself, as a pioneering feed-forward method, utilizes a transformer-based architecture to regress 3D point maps directly from image pairs. Spann3R~\cite{wang20243d} augments DUSt3R with a spatial memory network, allowing multiview inputs and improving efficiency by eliminating global alignment. However, Spann3R's sequential processing introduces error accumulation in reconstruction. Fast3R~\cite{yang2025fast3r} overcomes this limitation by introducing a global fusion transformer, processing multiple views simultaneously, and significantly improving reconstruction quality. In contrast, CUT3R~\cite{wang2025continuous} refines sequential reconstruction by maintaining and incrementally updating a persistent internal state that encodes scene content. 

Based on pointmap reconstruction, several methods have further developed high-quality novel view synthesis through 3D Gaussian reconstruction. Splatt3R~\cite{smart2024splatt3r} extends DUSt3R by adding a Gaussian head decoder that predicts Gaussian parameters directly from image pairs. LSM~\cite{fan2024large} similarly integrates a Gaussian head and further incorporates semantic embeddings from input images to augment anisotropic Gaussian predictions. NoPosplat~\cite{ye2024no}, after integrating a Gaussian head, performs full-parameter training to predict 3D Gaussians in a canonical space without relying on ground-truth camera poses or depth. PREF3R~\cite{chen2024pref3r}, based on Spann3R, also adds a Gaussian head to achieve 3D Gaussian predictions. SmileSplat~\cite{li2024smilesplat}, another Spann3R derivative, opts to predict Gaussian surfels instead of traditional 3D Gaussians. SelfSplat~\cite{kang2024selfsplat} integrates DUSt3R-based Gaussian predictions with self-supervised depth and pose estimation, jointly predicting depth, camera poses, and Gaussian attributes in a unified neural network. Lastly, FLARE~\cite{zhang2025flare} incorporates additional modules for pose estimation and global geometry projection, facilitating alignment of DUSt3R-based network token outputs.

Recent research has also explored pose-free feed-forward approaches at the object level. 
FORGE~\cite{jiang2024few} transforms per-view voxel features into a shared space using estimated relative camera poses and fuses them into a neural volume for rendering. 
LEAP~\cite{jiang2023leap} selects a canonical view from the input images, defines the neural volume in its local camera coordinate system, and reconstructs a radiance field by iteratively updating the volume via multiview encoding and a 2D-to-3D mapping module. 
PF-LRM~\cite{wang2023pf} jointly predicts a triplane NeRF and relative poses from sparse unposed images, supervising reconstruction with rendering losses and refining poses via a differentiable PnP solver.
MVDiffusion++\cite{tang2024mvdiffusion++} enables 3D consistency across views through 2D self-attention and view dropout, enabling dense and high-resolution synthesis without explicit pose supervision. 
SpaRP~\cite{xu2024sparp} pushes further by integrating sparse, unposed views into a composite image, which is then processed by a finetuned 2D diffusion model to enable both pose estimation and textured mesh reconstruction.

\subsection{Dynamic 3D Reconstruction \& View Synthesis}
Compared to the basic task of static scene reconstruction, dynamic scene reconstruction~\cite{tretschk2023state} poses significant challenges mainly due to the presence of moving objects, changing viewpoints, and temporal variations in scene geometry. 
Extending feedforward 3D reconstruction for dynamic scenarios mainly involves robust pose estimation to mitigate moving object interference, together with dynamic area segmentation for updating changing environments.

Seminal work on monocular depth estimation methods learned to predict temporally consistent depth video using temporal attention layers~\cite{cvd} and generative priors~\cite{hu2024depthcrafter, shao2024learning}. Though they demonstrate pleasure 3D points on camera space, they fail to provide global scene geometry due to the lack of camera pose estimation.

To jointly resolve pose and obtain a point cloud in canonical space, Robust-CVD~\cite{Robust-CVD} and CasualSAM~\cite{CasualSAM} integrate a depth prior with geometric optimization to estimate a smooth camera trajectory, as well as a detailed and stable depth and motion map reconstruction.
Most recently, MegaSaM~\cite{MegaSaM} further improves pose and depth accuracy by combining the strengths of several prior works, including DROID-SLAM~\cite{Droid}, optical flow~\cite{RAFT}, and a monocular depth estimation model~\cite{DepthAnything}, leading to results with previously unachievable quality.

Alternatively, instead of taking advantage of monocular prior models, some methods aim to train a dynamic 3D model from multiview 3D reconstruction models, \eg, DUSt3R~\cite{wang2024dust3r}.
MonST3R~\cite{zhang2025monst3r} estimates the pointmap at each timestep and processes them using a temporal sliding window to compute pairwise pointmap for each frame pair with MonST3R and optical flow from an off-the-shelf method. These intermediates then serve as inputs to optimize a global point cloud and per-frame camera poses and intrinsics. Video depth can be directly derived from this unified representation.
To speed up the optimization process in MonST3R, DAS3R~\cite{DAS3R} trains a dense prediction transformer~\cite{dpt} for motion segmentation inference and models the static scene as Gaussian splats with dynamics-aware optimization, allowing for more accurate background reconstruction results. 
%
Recent work CUT3R~\cite{wang2025continuous} fine-tunes DUSt3R on both static and dynamic data to enable feed-forward reconstruction, but it does not predict dynamic object segmentation. While effective, such approaches often require expensive training over a wide range of motion patterns to generalize well.  
%
In contrast, Easi3R~\cite{Easi3R} takes a different path, exploring a training-free and plug-and-play adaptation that enhances the generalization of DUSt3R variants for dynamic scene reconstruction, achieving accurate dynamic region segmentation, camera pose estimation, and 4D dense point map reconstruction at almost no additional cost on top of DUSt3R. 
To mitigate the conflict between camera pose estimation and geometry reconstruction requires,
PAGE-4D~\cite{zhou2025page} proposes to disentangle static and dynamic information by predicting a dynamic mask, suppressing motion cues for pose estimation while amplifying them for geometry reconstruction.
Driv3R~\cite{fei2024driv3r} further enables dynamic 3D reconstruction in large-scale autonomous driving scenarios by introducing a memory mechanism that supports efficient temporal integration. Besides, it also eliminates the global alignment optimization to reduce computational cost.

In addition to pointmap-based dynamic scene reconstruction, several recent methods based on 3D Gaussian Splatting (3DGS) have also been proposed for feed-forward dynamic reconstruction. L4GM \cite{ren2024l4gm} proposes the first 4D reconstruction model that produces animated objects from single-view videos using per-frame 3DGS representation. 4D-LRM~\cite{ma20254d} builds upon a transformer-based large reconstruction model, leveraging data-driven training for dynamic object reconstruction. It draws inspiration from 4D Gaussian Splatting~\cite{yang2023real} and reconstructs dynamic objects as anisotropic 4D Gaussian clouds. While prior works focus on dynamic object reconstruction, BulletTimer~\cite{liang2024feed} introduces the first feed-forward model for dynamic scene reconstruction. Building on GS-LRM~\cite{zhang2024gs}, it incorporates a bullet-time embedding into the input frames and aggregates information across all context frames, enabling feed-forward 3D Gaussian Splatting reconstruction at a specific timestamp. In addition, DGS-LRM~\cite{lin2025dgs} presents the first feed-forward deformable 3D Gaussian prediction from monocular videos using a transformer-based LRM, while 4DGT~\cite{xu20254dgt} extends transformer-based modeling to 4D Gaussian prediction from real-world monocular videos, demonstrating scalability in real-world settings.

Another line of research focuses on leveraging video pre-trained models for point map prediction by modeling 3D scenes as geometry videos. These approaches utilize diffusion models to learn the joint distribution of multiview RGB and geometric frames. A geometry video consists of standard RGB channels augmented with geometry channels, which encode structural information such as depth~\cite{Align3R}, XYZ coordinates~\cite{Sora3R}, color point rendering~\cite{Uni3C,ren2025gen3c}, or a combination of point-depth-ray maps~\cite{Geo4D}. 
Notably, Aether~\cite{aether} presents a unified framework that takes as input both image and action latents — such as ray maps — and produces predictions for images, actions, and depth. By flexibly combining different input conditions, Aether successfully achieved 4D dynamic reconstruction from video-only input, image-to-video generation from a single image, and camera-conditioned video synthesis given an image and a camera trajectory. 


To enable 3D point tracking, Stereo4D~\cite{Stereo4D} proposes a dynaDUSt3R architecture by incorporating a motion head for scene flow prediction. They use stereo videos from the Internet to create a dataset of more than 100,000 real-world 4D scenes with metric scale and long-term 3D motion trajectories for training.
Instead of predicting point map and flow map at reference and target viewpoints, St4RTrack~\cite{St4RTrack} outputs two point maps of different time steps for the reference view given two dynamic frames. The network is trained by reprojected supervision signals, including 2D trajectories and monocular depth, without the need for direct scene flow annotation. Inspired by ZeroCo~\cite{an2025cross}, D$^2$USt3R~\cite{han2025d} establishes dense correspondence of two pointmaps using the cross-attention maps of DUSt3R~\cite{wang2024dust3r}.

\subsection{Image Matching}

Recent advances in feed-forward 3D reconstruction have led to significant progress in image matching.
One notable example is MASt3R \cite{leroy2024grounding}, which builds on the DUSt3R \cite{wang2024dust3r} to enable efficient and robust image matching in a single forward pass. By augmenting the DUSt3R architecture with a dedicated head for dense local feature extraction, MASt3R introduces a mechanism to improve matching accuracy while maintaining the robustness characteristic of pointmap-based regression. However, MASt3R is fundamentally limited to processing image pairs with poor scalability for large image collections.
To address this issue, MASt3R-SfM \cite{duisterhof2024mast3r} proposes to leverage the frozen encoder of MASt3R for image retrieval, enabling it to process large and unconstrained image collections with quasi-linear complexity in a scalable way.
Importantly, the robustness of MASt3R’s local reconstructions allows the SfM pipeline to dispense with traditional RANSAC-based filtering. Instead, optimization is performed through successive gradient-based refinement in both 3D space (via a matching loss) and 2D image space (via reprojection loss), thus highlighting the potential of feed-forward paradigms to serve as both matching engines and geometric optimizers.

In parallel, to improve generalization for stereo matching, Wen et al. introduce FoundationStereo~\cite{wen2025foundationstereo}, a large-scale foundation model designed for zero-shot stereo correspondence estimation, pushing the state-of-the-art standard in stereo matching.

\subsection{3D-Aware Image Synthesis}
Feed-forward 3D models can be naturally applied to achieve 3D-aware image synthesis by combining 3D representation and generative models (e.g., GAN or diffusion models).

Several earlier methods employ voxel-based representations (e.g., PlatonicGAN~\cite{henzler2019escaping}) or 3D feature representations (e.g., HoloGAN~\cite{nguyen2019hologan} and BlockGAN~\cite{nguyen2020blockgan}). However, these approaches often suffer from limited multiview consistency. To address this, GRAF~\cite{schwarz2020graf} introduces 
a generative radiance field as a 3D representation, significantly improving consistency across different viewpoints. 
PiGAN~\cite{chan2021pi} leverages implicit neural representations with periodic activation functions to model scenes as view-consistent radiance fields. Subsequently, GIRAFFE~\cite{niemeyer2021giraffe} constructs compositional generative radiance fields for scene representations, enabling controllable image synthesis. StyleNeRF~\cite{gu2021stylenerf} combines NeRF-based 3D representations with a style-based generative model for high-resolution, 3D-consistent image synthesis. EG3D~\cite{chan2022efficient} introduces an explicit-implicit triplane representation to achieve efficient and high-quality 3D-aware image synthesis.
Due to the high computational cost of volume rendering in implicit NeRF-based scene representations, Hyun et al. propose GSGAN~\cite{hyun2024gsgan}, which replaces NeRF with 3D Gaussian Splatting (3DGS), enabling more efficient scene rendering through rasterization-based splatting. To stabilize the training of 3DGS-based 3D-aware image synthesis, GSGAN introduces hierarchical Gaussian representations, enabling coarse-to-fine scene modeling.

With the emergence of diffusion models, a series of works~\cite{chan2023generative,tewari2023diffusion,gu2023nerfdiff,szymanowicz2023viewset} make use of existing 2D diffusion backbones for 3D-aware synthesis by incorporating geometry priors in the form of 3D radiance fields. 
Instead of predicting NeRF, 
DiffSplat~\cite{lin2025diffsplat} proposes to generate 3D Gaussian parameters by fine-tuning image diffusion models with structured 3DGS	representations.
To mitigate the difficulty of 3DGS prediction,
Bolt3D~\cite{szymanowicz2025bolt3d} employs a diffusion model to predict pointmaps, which are further decoded into 3DGS parameters in a feed-forward manner.

\subsection{Camera-Controlled Video Generation} 
Camera-controlled video generation can be viewed as a natural extension of feed-forward 3D-free view synthesis into video generation.
To enable camera pose control in the video generation process, MotionCtrl~\cite{wang2024motionctrl}, CameraCtrl~\cite{he2024cameractrl}, I2VControl-Camera~\cite{feng2024i2vcontrol} inject the camera parameters (extrinsic, Plücker embedding, or point trajectory) into a pretrained video diffusion model. Building upon this, CamCo~\cite{xu2024camco} integrates epipolar constraints into attention layers, while CamTrol~\cite{hou2024training}, NVS-Solver~\cite{you2024nvs}, and ViewExtrapolator~\cite{liu2024novel} leverage explicit 3D point cloud renderings to guide the sampling process of the video diffusion models in a training-free manner. AC3D~\cite{bahmani2025ac3d} carefully designs the camera representation injection to the pretrained model. ViewCrafter~\cite{yu2024viewcrafter}, Gen3C~\cite{ren2025gen3c}, and See3D~\cite{ma2025you} fine-tuned video diffusion models on point cloud renderings to enable better novel view synthesis. VD3D~\cite{bahmani2024vd3d} enables camera control to transformer-based video diffusion models.
Beyond static scenes, CameraCtrl II~\cite{he2025cameractrl}, and ReCamMaster~\cite{bai2025recammaster} enable camera-controlled video generation on dynamic scenes by conditioning the video diffusion models on camera extrinsic parameters, while TrajectoryCrafter~\cite{yu2025trajectorycrafter} also enables dynamic scene view synthesis by conditioning the video diffusion models on dynamic point cloud. Several recent works have advanced beyond single-camera scenarios: CVD~\cite{kuang2024collaborative}, Vivid-ZOO~\cite{li2024vivid}, and SynCamMaster~\cite{bai2024syncammaster} develop frameworks for multi-camera synchronization.

\subsection{3D Understanding}
There have been works that embed features into feed-forward 3D reconstruction models, enabling 3D querying and segmentation through feature representations. Among earlier efforts, Large Spatial Model \cite{largespatialmodel} employs a point-based transformer that facilitates local context aggregation and hierarchical fusion to reconstruct a set of semantic anisotropic 3D Gaussians in a supervised end-to-end manner. GSemSplat \cite{gsemsplat} introduces a semantic head that predicts both region-specific and context-aware semantic features, which are then decoded into high-dimensional representations using MLP blocks for open-vocabulary semantic understanding. PE3R~\cite{hu2025pe3r} builds on the feed-forward pointmap method (e.g., DUSt3R) and a foundational segmentation model to achieve efficient semantic field reconstruction. In contrast to these three works, which focus on open-vocabulary segmentation, SplatTalk \cite{splattalk} tackles the broader challenge of free-form language reasoning required for 3D visual question answering (3D-VQA). It incorporates a feed-forward feature field as a submodule, including training a Gaussian encoder and a Gaussian latent decoder to reconstruct a 3D-language Gaussian field.

\renewcommand\arraystretch{1.15}
\begin{table*}[t]
\caption
{
Summarization of popular datasets for feed-forward 3D reconstruction and view synthesis.
}
\renewcommand\tabcolsep{4.2pt}
\centering
\scriptsize
\begin{tabular}{l || c c c c c c c c c c c c c}
\hline
\thickhline
\rowcolor{gray!20}

Datasets & \#Scenes (Objects) & Type & Real & Static & Dynamic & Camera & Point Cloud & Depth & Mesh & LiDAR & Semantic & Mask & Optical Flow\\

\hline\hline

\rowcolor{gray!10}
DTU~\cite{jensen2014large} & 124 & Objects
 & Real &  \textcolor{green}{\checkmark} & \textcolor{red}{\xmark} & \textcolor{green}{\checkmark} & \textcolor{green}{\checkmark} & \textcolor{red}{\xmark} & \textcolor{red}{\xmark} & \textcolor{red}{\xmark} & \textcolor{red}{\xmark} & \textcolor{red}{\xmark} & \textcolor{red}{\xmark}\\

Pix3D~\cite{sun2018pix3d} & 395 & Objects
 & Real &  \textcolor{green}{\checkmark} & \textcolor{red}{\xmark} & \textcolor{green}{\checkmark} & \textcolor{red}{\xmark} & \textcolor{red}{\xmark} & \textcolor{green}{\checkmark} & \textcolor{red}{\xmark} & \textcolor{red}{\xmark} & \textcolor{green}{\checkmark} & \textcolor{red}{\xmark}\\

\rowcolor{gray!10}
GSO~\cite{downs2022google} & 1,030 & Objects & Real & \textcolor{green}{\checkmark} & \textcolor{red}{\xmark} & \textcolor{green}{\checkmark} & \textcolor{red}{\xmark} & \textcolor{red}{\xmark} & \textcolor{green}{\checkmark} & \textcolor{red}{\xmark} & \textcolor{red}{\xmark} & \textcolor{red}{\xmark} & \textcolor{red}{\xmark}\\

OmniObject3D~\cite{wu2023omniobject3d} & 6,000 & Objects & Synthetic & \textcolor{green}{\checkmark} & \textcolor{red}{\xmark} & \textcolor{green}{\checkmark} & \textcolor{green}{\checkmark} & \textcolor{green}{\checkmark} & \textcolor{green}{\checkmark} & \textcolor{red}{\xmark} & \textcolor{red}{\xmark} & \textcolor{red}{\xmark} & \textcolor{red}{\xmark}\\

\rowcolor{gray!10}
CO3D~\cite{reizenstein2021common} & 18,619 & Objects & Real & \textcolor{green}{\checkmark} & \textcolor{red}{\xmark} & \textcolor{green}{\checkmark} & \textcolor{green}{\checkmark} & \textcolor{green}{\checkmark} & \textcolor{red}{\xmark} & \textcolor{red}{\xmark} & \textcolor{red}{\xmark} & \textcolor{green}{\checkmark} & \textcolor{red}{\xmark} \\

WildRGBD~\cite{xia2024rgbd} & 23,049 & Objects & Real & \textcolor{green}{\checkmark} & \textcolor{red}{\xmark} & \textcolor{green}{\checkmark} & \textcolor{green}{\checkmark} & \textcolor{green}{\checkmark} & \textcolor{red}{\xmark} & \textcolor{red}{\xmark} & \textcolor{red}{\xmark} & \textcolor{green}{\checkmark} & \textcolor{red}{\xmark} \\

\rowcolor{gray!10}
ShapeNet~\cite{chang2015shapenet} & 51,300 & Objects & Synthetic &  \textcolor{green}{\checkmark} &  \textcolor{red}{\xmark}  & \textcolor{red}{\xmark} & \textcolor{red}{\xmark} & \textcolor{red}{\xmark} & \textcolor{green}{\checkmark} & \textcolor{red}{\xmark} & \textcolor{red}{\xmark} & \textcolor{red}{\xmark} & \textcolor{red}{\xmark}\\

MVImgNet~\cite{yu2023mvimgnet} & 219,188 & Objects & Real & \textcolor{green}{\checkmark} & \textcolor{red}{\xmark} & \textcolor{green}{\checkmark} & \textcolor{green}{\checkmark} & \textcolor{red}{\xmark} & \textcolor{red}{\xmark} & \textcolor{red}{\xmark} & \textcolor{red}{\xmark} & \textcolor{green}{\checkmark} & \textcolor{red}{\xmark}\\

\rowcolor{gray!10}
Zeroverse~\cite{xie2024lrm} & 400K & Objects & Synthetic &  \textcolor{green}{\checkmark} &  \textcolor{red}{\xmark}  & \textcolor{red}{\xmark} & \textcolor{red}{\xmark} & \textcolor{red}{\xmark} & \textcolor{green}{\checkmark} & \textcolor{red}{\xmark} & \textcolor{red}{\xmark} & \textcolor{red}{\xmark} & \textcolor{red}{\xmark}\\

Objaverse~\cite{deitke2023objaverse} & 818K & Objects & Synthetic &  \textcolor{green}{\checkmark} &  \textcolor{green}{\checkmark}  & \textcolor{red}{\xmark} & \textcolor{red}{\xmark} & \textcolor{red}{\xmark} & \textcolor{green}{\checkmark} & \textcolor{red}{\xmark} & \textcolor{red}{\xmark} & \textcolor{red}{\xmark} & \textcolor{red}{\xmark}\\

\rowcolor{gray!10}
Objaverse-XL~\cite{deitke2023objaversexl} & 10.2M & Objects & Synthetic  &  \textcolor{green}{\checkmark} &  \textcolor{green}{\checkmark} & \textcolor{red}{\xmark} & \textcolor{red}{\xmark} & \textcolor{red}{\xmark} & \textcolor{green}{\checkmark} & \textcolor{red}{\xmark} & \textcolor{red}{\xmark} & \textcolor{red}{\xmark} & \textcolor{red}{\xmark}\\

7Scenes~\cite{shotton2013scene} & 7 & Indoor Scenes & Real &  \textcolor{green}{\checkmark} &  \textcolor{red}{\xmark}  & \textcolor{green}{\checkmark} & \textcolor{red}{\xmark} & \textcolor{green}{\checkmark} & \textcolor{green}{\checkmark} & \textcolor{red}{\xmark} & \textcolor{red}{\xmark} & \textcolor{red}{\xmark} & \textcolor{red}{\xmark}\\

\rowcolor{gray!10}
Replica~\cite{straub2019replica} & 18 & Indoor Scenes & Real  & \textcolor{green}{\checkmark} & \textcolor{red}{\xmark}  & \textcolor{red}{\xmark} & \textcolor{red}{\xmark} & \textcolor{red}{\xmark} & \textcolor{green}{\checkmark} & \textcolor{red}{\xmark} & \textcolor{green}{\checkmark}  & \textcolor{red}{\xmark} & \textcolor{red}{\xmark}\\

TUM RGBD~\cite{sturm2012evaluating} & 39 & Indoor Scenes & Real & \textcolor{green}{\checkmark} & \textcolor{green}{\checkmark} & \textcolor{green}{\checkmark} & \textcolor{red}{\xmark} & \textcolor{green}{\checkmark} & \textcolor{red}{\xmark} & \textcolor{red}{\xmark} & \textcolor{red}{\xmark} & \textcolor{red}{\xmark} & \textcolor{red}{\xmark}\\

\rowcolor{gray!10}
Matterport3D~\cite{chang2017matterport3d} & 90 & Indoor Scenes & Real  & \textcolor{green}{\checkmark} & \textcolor{red}{\xmark} & \textcolor{green}{\checkmark}  & \textcolor{red}{\xmark} & \textcolor{green}{\checkmark} & \textcolor{green}{\checkmark} & \textcolor{red}{\xmark} &  \textcolor{green}{\checkmark} & \textcolor{red}{\xmark} & \textcolor{red}{\xmark} \\

HyperSim~\cite{roberts2021hypersim} & 461 & Indoor Scenes & Synthetic & \textcolor{green}{\checkmark} & \textcolor{red}{\xmark} & \textcolor{green}{\checkmark} & \textcolor{red}{\xmark} & \textcolor{green}{\checkmark} & \textcolor{green}{\checkmark} & \textcolor{red}{\xmark} & \textcolor{green}{\checkmark} & \textcolor{red}{\xmark} & \textcolor{red}{\xmark} \\

\rowcolor{gray!10}
Dynamic Replica~\cite{karaev2023dynamicstereo} & 524 & Indoor Scenes & Synthetic  & \textcolor{red}{\xmark} & \textcolor{green}{\checkmark}  & \textcolor{green}{\checkmark} & \textcolor{red}{\xmark} & \textcolor{green}{\checkmark} & \textcolor{red}{\xmark} & \textcolor{red}{\xmark} & \textcolor{red}{\xmark}  & \textcolor{green}{\checkmark} & \textcolor{green}{\checkmark}\\

ScanNet++~\cite{yeshwanth2023scannet++} & 1,006 & Indoor Scenes & Real & \textcolor{green}{\checkmark} & \textcolor{red}{\xmark} & \textcolor{green}{\checkmark} & \textcolor{green}{\checkmark} & \textcolor{green}{\checkmark} & \textcolor{green}{\checkmark} & \textcolor{green}{\checkmark} & \textcolor{green}{\checkmark} & \textcolor{red}{\xmark} & \textcolor{red}{\xmark}\\

\rowcolor{gray!10}
ScanNet~\cite{dai2017scannet} & 1,513 & Indoor Scenes & Real &  \textcolor{green}{\checkmark} & \textcolor{red}{\xmark} & \textcolor{green}{\checkmark} & \textcolor{red}{\xmark} & \textcolor{green}{\checkmark} & \textcolor{green}{\checkmark} & \textcolor{red}{\xmark} & \textcolor{green}{\checkmark} & \textcolor{red}{\xmark} & \textcolor{red}{\xmark}\\

ARKitScenes~\cite{baruch2021arkitscenes} & 1,661 & Indoor Scenes & Real  & \textcolor{green}{\checkmark} & \textcolor{red}{\xmark} & \textcolor{green}{\checkmark} & \textcolor{green}{\checkmark} & \textcolor{green}{\checkmark} & \textcolor{green}{\checkmark} & \textcolor{green}{\checkmark} & \textcolor{red}{\xmark} & \textcolor{red}{\xmark} & \textcolor{red}{\xmark} \\

\rowcolor{gray!10}
MegaSynth~\cite{jiang2025megasynth} & 700K & Indoor Scenes & Synthetic & \textcolor{green}{\checkmark} & \textcolor{red}{\xmark} & \textcolor{green}{\checkmark} & \textcolor{red}{\xmark} & \textcolor{green}{\checkmark} & \textcolor{red}{\xmark} &  \textcolor{red}{\xmark} & \textcolor{red}{\xmark} & \textcolor{red}{\xmark} & \textcolor{red}{\xmark} \\

Virtual KITTI2~\cite{cabon2020virtual} & 5 & Outdoor Scenes & Synthetic & \textcolor{red}{\xmark} & \textcolor{green}{\checkmark} & \textcolor{green}{\checkmark} & \textcolor{red}{\xmark} & \textcolor{green}{\checkmark} & \textcolor{red}{\xmark} & \textcolor{red}{\xmark} & \textcolor{green}{\checkmark} & \textcolor{red}{\xmark} & \textcolor{green}{\checkmark} \\

\rowcolor{gray!10}
KITTI360~\cite{liao2022kitti} & 11 & Outdoor Scenes & Real & \textcolor{red}{\xmark} & \textcolor{green}{\checkmark} & \textcolor{green}{\checkmark} & \textcolor{green}{\checkmark} & \textcolor{red}{\xmark} & \textcolor{red}{\xmark} & \textcolor{green}{\checkmark} & \textcolor{green}{\checkmark} & \textcolor{red}{\xmark} & \textcolor{red}{\xmark} \\

Spring~\cite{mehl2023spring} & 47 &  Outdoor Scenes & Synthetic & \textcolor{red}{\xmark} & \textcolor{green}{\checkmark} & \textcolor{green}{\checkmark} & \textcolor{red}{\xmark} & \textcolor{green}{\checkmark} & \textcolor{red}{\xmark} & \textcolor{red}{\xmark} & \textcolor{red}{\xmark} & \textcolor{red}{\xmark} & \textcolor{green}{\checkmark} \\

\rowcolor{gray!10}
MegaDepth~\cite{li2018megadepth} & 196 &  Outdoor Scenes & Real & \textcolor{green}{\checkmark} & \textcolor{red}{\xmark} & \textcolor{green}{\checkmark} & \textcolor{red}{\xmark} & \textcolor{green}{\checkmark} & \textcolor{red}{\xmark} &  \textcolor{red}{\xmark} & \textcolor{red}{\xmark} & \textcolor{green}{\checkmark} & \textcolor{red}{\xmark} \\

ACID~\cite{liu2021infinite} &  13,047 & Outdoor Scenes & Real & \textcolor{green}{\checkmark} & \textcolor{red}{\xmark} & \textcolor{green}{\checkmark}  & \textcolor{red}{\xmark} & \textcolor{red}{\xmark} & \textcolor{red}{\xmark} & \textcolor{red}{\xmark} & \textcolor{red}{\xmark} & \textcolor{red}{\xmark} & \textcolor{red}{\xmark} \\

\rowcolor{gray!10}
MipNeRF360~\cite{barron2022mip} & 9 & \makecell[c]{Indoor \& Outdoor} & Real & \textcolor{green}{\checkmark} & \textcolor{red}{\xmark} & \textcolor{green}{\checkmark} & \textcolor{red}{\xmark} & \textcolor{red}{\xmark} & \textcolor{red}{\xmark} &  \textcolor{red}{\xmark} & \textcolor{red}{\xmark} & \textcolor{red}{\xmark} & \textcolor{red}{\xmark} \\

Tanks\&Temples~\cite{knapitsch2017tanks} & 21 & \makecell[c]{ Indoor \& Outdoor } & Real & \textcolor{green}{\checkmark} & \textcolor{red}{\xmark} & \textcolor{green}{\checkmark} & \textcolor{green}{\checkmark} & \textcolor{red}{\xmark} & \textcolor{green}{\checkmark} &  \textcolor{red}{\xmark} & \textcolor{red}{\xmark} & \textcolor{red}{\xmark} & \textcolor{red}{\xmark} \\

\rowcolor{gray!10}
ETH3D~\cite{schops2017multi} & 25 & \makecell[c]{ Indoor \& Outdoor } & Real &  \textcolor{green}{\checkmark} & \textcolor{red}{\xmark} & \textcolor{green}{\checkmark} & \textcolor{green}{\checkmark} & \textcolor{green}{\checkmark} & \textcolor{red}{\xmark} & \textcolor{red}{\xmark} & \textcolor{red}{\xmark} & \textcolor{green}{\checkmark} & \textcolor{red}{\xmark}\\

PointOdyssey~\cite{zheng2023pointodyssey} & 159 & \makecell[c]{ Indoor \& Outdoor } & Synthetic & \textcolor{red}{\xmark} & \textcolor{green}{\checkmark} & \textcolor{green}{\checkmark} & \textcolor{red}{\xmark} & \textcolor{green}{\checkmark} & \textcolor{red}{\xmark} & \textcolor{red}{\xmark} & \textcolor{red}{\xmark} & \textcolor{green}{\checkmark} & \textcolor{red}{\xmark} \\

\rowcolor{gray!10}
TartanAir~\cite{wang2020tartanair} & 1,037 & \makecell[c]{ Indoor \& Outdoor } & Synthetic & \textcolor{green}{\checkmark} & \textcolor{green}{\checkmark} & \textcolor{green}{\checkmark} & \textcolor{green}{\checkmark} & \textcolor{green}{\checkmark} & \textcolor{red}{\xmark} & \textcolor{green}{\checkmark} & \textcolor{green}{\checkmark} & \textcolor{red}{\xmark} & \textcolor{green}{\checkmark} \\

DL3DV-10K~\cite{ling2024dl3dv} & 10,510 & \makecell[c]{ Indoor \& Outdoor } & Real  & \textcolor{green}{\checkmark} & \textcolor{red}{\xmark} & \textcolor{green}{\checkmark} & \textcolor{red}{\xmark} & \textcolor{red}{\xmark}  & \textcolor{red}{\xmark} & \textcolor{red}{\xmark} & \textcolor{red}{\xmark} & \textcolor{red}{\xmark} & \textcolor{red}{\xmark} \\

\rowcolor{gray!10}
RealEstate10K~\cite{zhou2018stereo} & 74,766 & \makecell[c]{ Indoor \& Outdoor } & Real & \textcolor{green}{\checkmark} & \textcolor{red}{\xmark} & \textcolor{green}{\checkmark} & \textcolor{red}{\xmark} & \textcolor{red}{\xmark} & \textcolor{red}{\xmark} &  \textcolor{red}{\xmark} & \textcolor{red}{\xmark} & \textcolor{red}{\xmark} & \textcolor{red}{\xmark} \\

BlendedMVS~\cite{deitke2023objaverse} & 113 & \makecell[c]{ Indoor \& Outdoor } & Synthetic &  \textcolor{green}{\checkmark} &  \textcolor{red}{\xmark}  & \textcolor{green}{\checkmark} & \textcolor{red}{\xmark} & \textcolor{green}{\checkmark} & \textcolor{green}{\checkmark} & \textcolor{red}{\xmark} & \textcolor{red}{\xmark} & \textcolor{green}{\checkmark} & \textcolor{red}{\xmark}\\


\end{tabular}
\label{tab_datasets}
\end{table*}

\subsection{Digital Human}
Recent progress in feed-forward 3D reconstruction has attracted increasing attention in photorealistic 3D avatars. For example, GPS-Gaussian~\cite{zheng2024gps} defines 2D Gaussian parameter maps on the input views and directly predicts 3D Gaussians in a feed-forward manner, enabling efficient and generalizable human novel view synthesis. Avat3r~\cite{kirschstein2025avat3r} builds upon the Large Gaussian Reconstruction Model~\cite{xu2024grm} to predict 3D Gaussians corresponding to each pixel of the input image, achieving animatable 3D reconstruction and high-quality 3D head avatars. Additionally, GaussianHeads~\cite{teotia2024gaussianheads} and GIGA~\cite{zubekhin2025giga} further enhance 3DGS-based avatars, with the former enabling real-time head reconstruction under large deformations and the latter achieving large-scale generalization by training on thousands of data. However, only a few works is superficially covered as digital human is not the primary focus of this survey.

\subsection{SLAM \& Visual Localization}
Recent SLAM systems have increasingly adopted feed-forward models to replace traditional geometric pipelines, offering real-time and dense reconstruction from monocular RGB videos.
MASt3R-SLAM~\cite{murai2025mast3r} leverages the MASt3R~\cite{leroy2024grounding} prior to build a real-time dense monocular SLAM system that operates without requiring known camera calibration. 
Similarly, based on DUSt3R, SLAM3R~\cite{liu2025slam3r} introduces a real-time, end-to-end dense reconstruction system that directly predicts 3D pointmaps from RGB videos. Its Image-to-Points (I2P) module extends DUSt3R to multiview inputs for improved local geometry, while the Local-to-World (L2W) module incrementally aligns local pointmaps into a global frame—eliminating the need for camera pose estimation or global optimization.
However, MASt3R and DUSt3R, being inherently two-view, limit each inference to a fixed image pair, making large-scale fusion dependent on iterative matching and optimization. VGGT-SLAM \cite{maggio2025vggt} addresses this limitation by adopting the more powerful VGGT transformer, which supports arbitrary-length image sets (within memory constraints) and jointly predicts dense point clouds, camera poses, and intrinsics in a single forward pass. This allows VGGT-SLAM to construct larger submaps and align them via projective transformations optimized on the SL(4) manifold. 

For visual localization, Reloc3R~\cite{dong2025reloc3r} builds on DUSt3R as its backbone and introduces a symmetric relative pose regression and a motion averaging module, enabling strong generalization with accurate camera pose estimation.

\subsection{Robot Manipulation}
GraspNeRF~\cite{graspnerf} employs a generalizable NeRF to predict TSDF values, and then a grasp prediction network takes TSDF values as input to predict grasping poses for transparent and specular objects. ManiGaussian~\cite{manigaussian} adopts a feed-forward 3DGS model for robotics manipulation. It introduces a dynamic GS framework to model the propagation of diverse semantic features, along with a Gaussian world model that supervises learning by reconstructing future scenes for scene-level dynamics mining. Its follow-up work ManiGaussian++ \cite{manigaussian++}, extends
ManiGaussian by introducing the hierarchical Gaussian
world model to learn the multibody spatiotemporal
dynamics for bimanual tasks. While many works use optimization-based NeRF and 3D Gaussians for robotics tasks like manipulation and navigation, few adopt feed-forward 3D models due to reconstruction quality concerns. However, as feed-forward reconstruction quality rapidly improves, more works are expected to shift toward these models for their significantly faster inference speed.

\begin{table*}[t]
\small
\centering
\scriptsize
\setlength\tabcolsep{2.1pt}
\renewcommand\arraystretch{1.1}
\begin{tabular}{l||ccc|ccc|ccc||ccc|ccc}
\hline
\thickhline
\rowcolor{gray!20}
 & 
\multicolumn{3}{c|}{\textbf{Sintel}} &  
\multicolumn{3}{c|}{\textbf{TUM-dynamics}} & 
\multicolumn{3}{c||}{\textbf{ScanNet}} &
\multicolumn{3}{c|}{\textbf{RealEstate10K}} &
\multicolumn{3}{c}{\textbf{Co3Dv2}}
\\
\rowcolor{gray!20}
\multirow{-2}{*}{\textbf{Methods}} &
ATE$\downarrow$ & RPE trans$\downarrow$ & RPE rot$\downarrow$ &
        ATE$\downarrow$ & RPE trans$\downarrow$ & RPE rot$\downarrow$ &
        ATE$\downarrow$ & RPE trans$\downarrow$ & RPE rot$\downarrow$ &
        RRA@30 $\uparrow$ & RTA@30 $\uparrow$ & AUC@30 $\uparrow$ &
        RRA@30 $\uparrow$ & RTA@30 $\uparrow$ & AUC@30 $\uparrow$
\\
\hline\hline
\rowcolor{gray!10}
DUSt3R~\cite{wang2024dust3r} & 0.290 & 0.132 & 7.869 & 0.140 & 0.106 & 3.286 & 0.246 & 0.108 & 8.210 & - & - & - & - & - & - \\

Fast3R~\cite{yang2025fast3r} & 0.371 & 0.298 & 13.75 & 0.090 & 0.101 & 1.425 & 0.155 & 0.123 & 3.491 & 99.05 & 81.86 & 61.68 & 97.49 & 91.11 & 73.43\\

\rowcolor{gray!10}
Spann3R~\cite{wang20243d} & 0.329 & 0.110 & 4.471 & 0.056 & 0.021 & 0.591 & 0.096 & 0.023 & 0.661
& - & - & - & - & - & - \\

CUT3R~\cite{wang2025continuous} & 0.217 & 0.070 & 0.636 & 0.047 & 0.015 & 0.451 & 0.094 & 0.022 & 0.629 & 99.82 & 95.10 & \underline{81.47} & 96.19 & 92.69 & 75.82 \\

\rowcolor{gray!10}
STream3R~\cite{lan2025stream3r} & 0.213 & 0.076 & 0.868 & 0.026 & 0.013 & 0.330 & 0.052 & 0.021 & 0.850
 & - & - & - & - & - & - \\

Aether~\cite{aether} & 0.189 & 0.054 & 0.694 & 0.092 & 0.012 & 1.106 & 0.176 & 0.028 & 1.204 & - & - & - & - & - & - \\

\rowcolor{gray!10}
MonST3R~\cite{zhang2025monst3r} & \underline{0.108} & \underline{0.042} & 0.732 & 0.074 & 0.019 & 0.905 & 0.068 & 0.017 & 0.545 
& - & - & - & - & - & - \\

FLARE~\cite{zhang2025flare} & 0.207 & 0.090 & 3.015 & 0.026 & 0.013 & 0.475 & 0.064 & 0.023 & 0.971 & 99.69 & \underline{95.23} & 80.01 & 96.38 & 93.76 & 73.99 \\

\rowcolor{gray!10}
VGGT~\cite{wang2025vggt} & 0.167 & 0.062 & \underline{0.491} & \textbf{0.012} & \underline{0.010} & \textbf{0.311} & \underline{0.035} & \underline{0.015} & \underline{0.382}  &   \underline{99.97} & 93.13 & 77.62 & \underline{98.96} & \underline{97.13} & \textbf{88.59} \\

PAGE-4D~\cite{zhou2025page} & 0.143 & 0.078 & 0.538 & 0.016 & 0.011 & 0.323
& - & - & - & - & - & - & - & - & - \\

\rowcolor{gray!10}
$\pi^3$~\cite{wang2025pi} & \textbf{0.074} & \textbf{0.040} & \textbf{0.282} & \underline{0.014} & \textbf{0.009} & \underline{0.312} & \textbf{0.031} & \textbf{0.013} & \textbf{0.347}       & \textbf{99.99} & \textbf{95.62} & \textbf{85.90} & \textbf{99.05} & \textbf{97.33} & \underline{88.41}  \\

\end{tabular}
\caption{
Evaluation of camera pose estimation. 
For Sintel~\cite{bozic2021transformerfusion}, TUMdynamics~\cite{sturm2012benchmark}, and ScanNet~\cite{dai2017scannet}, metrics measure the distance error of rotation/translation. 
For RealEstate10K~\cite{zhou2018stereo} and Co3Dv2~\cite{reizenstein2021common}, metrics measure the distance error of rotation/translation. 
Results are adopted from \cite{wang2025pi,zhou2025page,lan2025stream3r}.
}
\label{tab:camera_pose}
\end{table*}

\begin{table*}[t]
\centering
\scriptsize
\setlength\tabcolsep{7.5pt}
\renewcommand\arraystretch{1.1}
\begin{tabular}{lc||cccccc|cccccc}
\hline
\thickhline

\rowcolor{gray!20}
        & &
        \multicolumn{6}{c|}{\textbf{7-Scenes}} &
        \multicolumn{6}{c}{\textbf{NRGBD}} \\
        \cline{3-14}  
\rowcolor{gray!20}
        & &
        \multicolumn{2}{c}{Accuracy $\downarrow$}  &
        \multicolumn{2}{c}{Completion $\downarrow$} &
        \multicolumn{2}{c|}{Normal Consistency $\uparrow$}     & 
        \multicolumn{2}{c}{Accuracy $\downarrow$}  &
        \multicolumn{2}{c}{Completion $\downarrow$} &
        \multicolumn{2}{c}{Normal Consistency $\uparrow$}     \\

        \cline{3-14}
\rowcolor{gray!20}
\multirow{-3}{*}{\textbf{Methods}} & \multirow{-3}{*}{\textbf{Types}} &
        Mean & Median &
        Mean & Median &
        Mean & Median &
        Mean & Median &
        Mean & Median &
        Mean & Median  \\

\hline\hline
\rowcolor{gray!10}
DUSt3R-GA~\cite{wang2024dust3r} & Optim. & 0.146 & 0.077 & 0.181 & 0.067 & 0.736 & 0.839 & 0.144 & 0.019 & 0.154 & 0.018 & 0.870 & 0.982 \\

MASt3R-GA~\cite{murai2025mast3r} & Optim. & 0.185 & 0.081 & 0.180 & 0.069 & 0.701 & 0.792 & 0.085 & 0.033 & 0.063 & 0.028 & 0.794 & 0.928  \\

\rowcolor{gray!10}
MonST3R-GA~\cite{zhang2025monst3r} & Optim. & 0.248 & 0.185 & 0.266 & 0.167 & 0.672 & 0.759 & 0.272 & 0.114 & 0.287 & 0.110 & 0.758 & 0.843 \\

Spann3R~\cite{wang20243d} & Stream & 0.298 & 0.226 & 0.205 & 0.112 & 0.650 & 0.730 & 0.416 & 0.323 & 0.417 & 0.285 & 0.684 & 0.789 \\

\rowcolor{gray!10}
CUT3R~\cite{wang2025continuous} & Stream & 0.126 & 0.047 & 0.154 & 0.031 & 0.727 & 0.834 & 0.099 & 0.031 & 0.076 & 0.026 & 0.837 & 0.971 \\

STream3R~\cite{lan2025stream3r} & Stream & 0.122 & 0.044 & 0.110 & 0.038 & 0.746 & 0.856 
 & - & - & - & - & - & -  \\

\rowcolor{gray!10}
SLAM3R~\cite{liu2025slam3r} & Stream & 0.287 & 0.155 & 0.226 & 0.066 & 0.644 & 0.720
 & - & - & - & - & - & -  \\

Fast3R~\cite{yang2025fast3r} & Full & 0.164 & 0.108 & 0.163 & 0.080 & 0.686 & 0.775
 & - & - & - & - & - & -  \\

\rowcolor{gray!10}
VGGT~\cite{wang2025vggt} & Full & 0.087 & 0.039 & 0.091 & 0.039 & 0.787 & 0.890 & - & - & - & - & - & -  \\

\end{tabular}
\caption{
Evaluation of point map estimation on 7-Scenes~\cite{shotton2013scene} and NRGBD~\cite{azinovic2022neural}. `GA' denotes Global Alignment.
Results are mainly adopted from \cite{wang2025continuous,lan2025stream3r}.
}
\label{tab:pointmap}
\end{table*}

\begin{table*}[t]
\centering
\scriptsize
\setlength\tabcolsep{5pt}
\renewcommand\arraystretch{1.1}
\begin{tabular}{l||cccccc|cccccc}
\hline
\thickhline

\rowcolor{gray!20}
        &
        \multicolumn{6}{c|}{\textbf{Scale Alignment}} &
        \multicolumn{6}{c}{\textbf{Scale \& Shift Alignment}} \\
        \cline{2-13}  
\rowcolor{gray!20}
        &
        \multicolumn{2}{c}{Sintel}  &
        \multicolumn{2}{c}{Bonn} &
        \multicolumn{2}{c|}{KITTI}     & 
        \multicolumn{2}{c}{Sintel}  &
        \multicolumn{2}{c}{Bonn} &
        \multicolumn{2}{c}{KITTI}     \\

        \cline{2-13}
\rowcolor{gray!20}
\multirow{-3}{*}{\textbf{Methods}} & 

        Abs Rel $\downarrow$ & $\delta<1.25$ $\uparrow$ &
        Abs Rel $\downarrow$ & $\delta<1.25$ $\uparrow$ &
        Abs Rel $\downarrow$ & $\delta<1.25$ $\uparrow$ & 
        Abs Rel $\downarrow$ & $\delta<1.25$ $\uparrow$ &
        Abs Rel $\downarrow$ & $\delta<1.25$ $\uparrow$ &
        Abs Rel $\downarrow$ & $\delta<1.25$ $\uparrow$ \\

\hline\hline
\rowcolor{gray!10}
DUSt3R~\cite{wang2024dust3r} & 0.662 & 0.434 & 0.151 & 0.839 & 0.143 & 0.814 & 0.570 & 0.493 & 0.152 & 0.835 & 0.135 & 0.818 \\

MASt3R~\cite{murai2025mast3r} & 0.558 & 0.487 & 0.188 & 0.765 & 0.115 & 0.848 & 0.480 & 0.517 & 0.189 & 0.771 & 0.115 & 0.849 \\

\rowcolor{gray!10}
MonST3R~\cite{zhang2025monst3r} & 0.399 & 0.519 & 0.072 & 0.957 & 0.107 & 0.884 & 0.402 & 0.526 & 0.070 & 0.958 & 0.098 & 0.883 \\

Fast3R~\cite{yang2025fast3r} & 0.638 & 0.422 & 0.194 & 0.772 & 0.138 & 0.834 & 0.518 & 0.486 & 0.196 & 0.768 & 0.139 & 0.808 \\

\rowcolor{gray!10}
MVDUSt3R+~\cite{tang2025mv} & 0.805 & 0.283 & 0.426 & 0.357 & 0.456 & 0.342 & 0.619 & 0.332 & 0.482 & 0.357 & 0.401 & 0.355 \\
CUT3R~\cite{wang2025continuous} & 0.417 & 0.507 & 0.078 & 0.937 & 0.122 & 0.876 & 0.534 & 0.558 & 0.075 & 0.943 & 0.111 & 0.883 \\

\rowcolor{gray!10}
Stream3R~\cite{lan2025stream3r} & \underline{0.264} & \textbf{70.5} & 0.069 & 0.952 & 0.080 & 0.947 & - & - & - & - & - & - \\
Aether~\cite{aether} & 0.324 & 0.502 & 0.273 & 0.594 & \underline{0.056} & \underline{0.978} & 0.314 & 0.604 & 0.308 & 0.602 & 0.054 & \underline{0.977} \\

\rowcolor{gray!10}
FLARE~\cite{zhang2025flare} & 0.729 & 0.336 & 0.152 & 0.790 & 0.356 & 0.570 & 0.791 & 0.358 & 0.142 & 0.797 & 0.357 & 0.579 \\
VGGT~\cite{wang2025vggt} & 0.299 & \underline{0.638} & \underline{0.057} & \underline{0.966} & 0.062 & 0.969 & \underline{0.230} & \underline{0.678} & \underline{0.052} & \underline{0.969} & \underline{0.052} & 0.968 \\

\rowcolor{gray!10}
\textbf{$\pi^3$}~\cite{wang2025pi} & \textbf{0.233} & \underline{0.664} & \textbf{0.049} & \textbf{0.975} & \textbf{0.038} & \textbf{0.986} & \textbf{0.210} & \textbf{0.726} & \textbf{0.043} & \textbf{0.975} & \textbf{0.037} & \textbf{0.985} \\

\end{tabular}
\caption{
Evaluation of video depth estimation on Sintel~\cite{bozic2021transformerfusion}, Bonn~\cite{palazzolo2019refusion} and KITTI~\cite{geiger2013vision}. Results are mainly adopted from \cite{wang2025pi}.
}
\label{tab:video_depth}
\end{table*}

\begin{table*}[t]
\centering
\scriptsize
\setlength\tabcolsep{13.25pt}
\renewcommand\arraystretch{1.1}
\begin{tabular}{l||ccc|ccc|ccc}
\hline
\thickhline
\rowcolor{gray!20}
 & 
\multicolumn{3}{c|}{\textbf{Tanks-and-Temples}} &  
\multicolumn{3}{c|}{\textbf{RealEstate10K}} & 
\multicolumn{3}{c}{\textbf{DL3DV}}
\\
\cline{2-10} 
\rowcolor{gray!20}
\multirow{-2}{*}{\textbf{Methods}} &
        PSNR$\uparrow$ & SSIM$\uparrow$ & LPIPS$\downarrow$ &
        PSNR$\uparrow$ & SSIM$\uparrow$ & LPIPS$\downarrow$ &
        PSNR$\uparrow$ & SSIM$\uparrow$ & LPIPS$\downarrow$ 
\\

\hline\hline

\rowcolor{gray!10}
ZeroNVS \cite{sargent2024zeronvs}       & 13.14 & 0.327 & 0.516  & 15.23 & 0.540  & 0.386 & 14.17 & 0.441  & 0.481 \\

CameraCtrl \cite{he2024cameractrl}       & 15.34 & 0.534 & 0.331 & 17.74 & 0.681 & 0.278 & 16.31 & 0.552 & 0.352 \\

\rowcolor{gray!10}
GenWarp \cite{seo2024genwarp}          & 16.45 & 0.513 & 0.377 & 15.30 & 0.538 & 0.371 & 15.81 & 0.531 & 0.382 \\

ViewCrafter \cite{yu2024viewcrafter}       & 17.18 & 0.589 & 0.346  & 17.75 & 0.681  & 0.315 & 17.24 & 0.571  &0.329 \\

\rowcolor{gray!10}
DimensionX \cite{sun2024dimensionx}  & 17.78 & 0.635 & 0.228     & 18.21 & 0.717  & 0.307 & 18.22 & 0.653  & 0.201 \\ 

SEVA \cite{zhou2025stable}   & 17.61 & 0.621 & 0.235    & 17.58 & 0.688  & 0.334 & 18.01 & 0.638  & 0.214 \\

\rowcolor{gray!10}
MVGenMaster \cite{cao2025mvgenmaster}  & 18.03 & 0.622 & 0.253 & 17.87 & 0.701 & 0.321 & 17.71 & 0.586 & 0.277\\

See3D \cite{ma2025you}    &  18.35 & 0.641 & 0.244  & 18.24 & 0.735 & 0.293 & 18.41 & 0.631 & 0.215 \\

\rowcolor{gray!10}
Voyager \cite{huang2025voyager} & 18.61 & 0.669 & 0.238 & 18.56 & 0.723 & 0.264 & 18.84 & 0.636  & 0.227 \\

GEN3C \cite{ren2025gen3c}      & 19.18 & 0.681 & 0.207  & 20.64 & 0.754  & 0.229 & 19.14 & 0.658  & 0.198 \\

\rowcolor{gray!10}
PE-Fields \cite{bai2025positional}  & 22.12 & 0.732 & 0.174 & 21.65 & 0.816 & 0.162 & 22.23 & 0.742 & 0.154

\end{tabular}
\caption{
Evaluation of novel view synthesis from a single input image on Tanks-and-Temples~\cite{knapitsch2017tanks}, RealEstate10K~\cite{zhou2018stereo}, and DL3DV~\cite{ling2024dl3dv} datasets. Results are mainly adopted from \cite{bai2025positional}.
}
\label{tab:nvs}
\end{table*}

\section{Evaluation}
\label{experiments}

\subsection{Datasets}

Datasets are the core of feed-forward 3D reconstruction and view synthesis. To give an overall picture of the datasets, we tabulate detailed scene and annotation types in popular datasets in Table~\ref{tab_datasets}. The scene types are divided into objects, indoor and outdoor scenes. And we also indicate synthetic datasets (e.g., Objaverse~\cite{deitke2023objaverse}), where MegaSynth~\cite{jiang2025megasynth} and Zeroverse~\cite{xie2024lrm} are procedurally synthesized datasets, real-world datasets (e.g., ACID~\cite{liu2021infinite}), static datasets (e.g., ARKitScenes~\cite{baruch2021arkitscenes}) and dynamic datasets (e.g., KITTI360~\cite{liao2022kitti}). Notably, several datasets, for example TartanAir~\cite{wang2020tartanair}, include both static and dynamic scenes.

\subsection{Evaluation Metrics}

Several metrics have been widely adopted for faithful evaluations in various feed-forward 3D reconstruction and view synthesis tasks. For novel view synthesis evaluation, PSNR (Peak Signal-to-Noise Ratio), SSIM (Structural Similarity Index) \cite{wang2004image}, and LPIPS (Learned Perceptual Image Patch Similarity) \cite{LPIPS} are commonly used to evaluate image quality from different perspectives. 

For camera pose estimation, RTA (Relative Translation Accuracy), RRA (Relative Rotation Accuracy), and AUC (Area Under Curve) are widely adopted. RTA and RRA measure the relative angular errors in translation and rotation between image pairs, respectively. AUC computes the area under the accuracy curve across different angular thresholds.
In point map evaluation, standard metrics include point cloud Accuracy (or precision), Completeness (or recall), and Chamfer distance. The point cloud accuracy is the average nearest-neighbor distance from each predicted point to the ground-truth surface, indicating how precisely predicted points are placed. The point cloud completeness is the average nearest-neighbor distance from each ground-truth point to the reconstruction, reflecting how fully the ground-truth surface is covered. The Chamfer Distance combines the Accuracy and Completeness scores.

For monocular depth estimation, people usually calculate the absolute relative error $|y - \hat{y}|/y$ where $y$ is the ground-truth and $\hat{y}$ is the prediction, and the percentage of inlier points $\delta < T$, which is the percentage of predicted depths within a T-factor of true depth. 
For multiview depth estimation, several metrics are usually reported including:
1) Accuracy, which measures the smallest Euclidean distance from the prediction to the ground-truth surface;
2) Completeness, which measures the smallest Euclidean distance from the ground truth to prediction;
3) Overall, which is the mean of Accuracy and Completeness, equivalent to the Chamfer distance.
Notably, scale ambiguity is an unavoidable issue for monocular depth estimation. A common practice is to perform scale alignment during evaluation, such as median scaling or least-squares fitting.

For dynamic point tracking,  OA (Occlusion Accuracy), $\sigma^{vis}_{avg}$, and AJ (Average Jaccard) \cite{doersch2022tap} are used together. OA measures the binary accuracy of the occlusion predictions; $\sigma^{x}_{avg}$ measures the fraction of points that are accurately tracked within a certain pixel threshold;  Average Jaccard considers both occlusion and prediction accuracy.


\subsection{Evaluation Results}

Tables~\ref{tab:camera_pose}-\ref{tab:nvs} summarize quantitative results across major benchmarks, 
covering camera pose estimation, point map reconstruction, video depth estimation, 
and novel view synthesis. These comparisons demonstrate the rapid evolution of 
feed-forward 3D reconstruction models in terms of accuracy and scalability.

\textbf{Camera Pose Estimation.}
As shown in Table~\ref{tab:camera_pose}, recent transformer-based methods (e.g., \textbf{VGGT}~\cite{wang2025vggt} and \textbf{$\pi^3$}~\cite{wang2025pi}) achieve state-of-the-art performance on Sintel, TUM-dynamics, ScanNet, 
RealEstate10K, and Co3Dv2. Compared to early baselines such as DUSt3R and Spann3R, these large-scale models reduce both translation and rotation errors by more than 50\% while maintaining high accuracy under wide-baseline conditions. Notably, PAGE-4D~\cite{zhou2025page} improves pose estimation in dynamic scenes by disentangling static and dynamic cues, demonstrating strong robustness for real-world motion sequences.

\textbf{Point Map Estimation.}
Table~\ref{tab:pointmap} reports evaluations on 7-Scenes and NRGBD datasets. 
VGGT achieves the best overall results, attaining mean accuracy below 0.09 and the highest normal consistency (0.89), confirming the effectiveness of large transformer architectures in maintaining geometric fidelity.
Feed-forward stream-based approaches such as CUT3R and STream3R surpass optimization-based pipelines (DUSt3R-GA, MASt3R-GA) in both accuracy and normal consistency, without causing a large burden with the continuous updating mechanisms. 

\textbf{Video Depth Estimation.}
In Table~\ref{tab:video_depth}, models are evaluated on Sintel, Bonn, and KITTI. 
Under both scale-aligned and scale-shift aligned protocols, \textbf{$\pi^3$} again delivers the most accurate depth with the lowest absolute relative error (0.21–0.23) and highest inlier ratio ($\delta < 1.25$ up to 0.99). VGGT and CUT3R follow closely, while MVDUSt3R+ demonstrates stable performance 
across varying numbers of input views. These results suggest that feed-forward models 
can generalize to complex video geometry without per-sequence optimization.

\textbf{Novel View Synthesis.}
For single-image novel view synthesis (Table~\ref{tab:nvs}), generative diffusion-based approaches have advanced image quality on Tanks-and-Temples, RealEstate10K, and DL3DV datasets. 
\textbf{PE-Fields} achieves the highest PSNR (22.1–22.2) and lowest LPIPS (0.15–0.17), surpassing previous state-of-the-art methods such as SEVA and GEN3C.
Recent large multiview diffusion models (e.g., See3D and MVGenMaster) show consistent improvement in perceptual realism and structural similarity, 
marking a clear trend toward unified feed-forward novel view synthesis and image generation.


\section{Open Challenges}
\label{future}

Though feed-forward 3D models have made notable progress and achieved superior performance in recent years, there exist several challenges that need further exploration.
In this section, we provide an overview of typical challenges, share our humble opinions on possible solutions, and highlight future research directions.

\subsection{Limited Modality in Datasets}
Most existing 3D reconstruction and view synthesis datasets have a limited coverage of data modalities. Specifically, many widely-used benchmarks, such as RealEstate10K~\cite{zhou2018stereo} and MipNeRF360~\cite{barron2022mip}, comprise RGB images only without including essential complementary signals like depth, LiDAR, or semantic annotations. Even large-scale collections like Objaverse-XL~\cite{deitke2023objaversexl} (10.2M objects) focus primarily on synthetic mesh data, lacking the real-world data modalities needed to train robust models. Many studies address this imbalance issue by merging multiple datasets of different modalities, but this inevitably introduces domain shifts and annotation inconsistencies.
The modality limitation is particularly acute in the area of dynamic scene understanding. While several datasets provide dynamic sequences, those with comprehensive multi-modal annotations (e.g., synchronized RGB, depth, optical flow, and 3D motion) remain significantly fewer than their static counterparts. Most dynamic datasets prioritize either camera motion or object movement, but rarely capture both simultaneously with full sensor suites. This scarcity of richly annotated dynamic data severely constrains the development of models capable of handling real-world scenarios that often involve both camera motion and object motion.

A fundamental challenge emerges: how to create scalable, modality-rich datasets that combine the diversity of synthetic collections like Objaverse~\cite{deitke2023objaverse} with the multi-sensor completeness of real-world benchmarks such as ScanNet++~\cite{yeshwanth2023scannet++}. Current approaches address this issue by patching together incompatible data sources, which ultimately limits progress toward generalizable 3D understanding. The field is facing an urgent need of comprehensive resources that provide aligned multi-modal signals, including RGB, depth, and semantics, all collected under a unified protocol for mitigating the data modality limitation.

\subsection{Reconstruction Accuracy}

Feed-forward 3D reconstruction models have made notable progress in recent years. However, their reconstruction accuracy, particularly in terms of depth map precision, is still inferior to traditional multiview stereo (MVS) methods~\cite{goesele2006multi, yao2018mvsnet, zhang2023geomvsnet} that explicitly utilize camera parameters for all input frames. Specifically, MVS approaches typically leverage known camera parameters and hypothesized depth sets to construct cost volumes, subsequently processed to predict accurate depth or disparity maps.
An intriguing hypothesis is that feed-forward 3D reconstruction models might spontaneously learn an approximation of such cost volumes. Modern feed-forward reconstruction models~\cite{wang2025vggt, yang2025fast3r} mostly employ self-attention layers, theoretically enabling them to approximate or even exceed the representational capacity of traditional cost volumes. With sufficient high-quality training data, these feed-forward models have the potential to match and even surpass the accuracy of MVS-based methods.
Moreover, incorporating explicit camera parameters or additional priors into the feature backbone, such as through Diffusion Transformers~\cite{peebles2023scalable}, offers another promising avenue to improve the reconstruction accuracy. Consequently, we anticipate that feed-forward models will continue to evolve, eventually much outperforming traditional MVS methods and achieving sensor-level accuracy, comparable to technologies like LiDAR or high-precision scanning systems.

\subsection{Free-viewpoint Rendering}
The challenge of free-viewpoint rendering lies in the difficulty of generating high-quality novel views that are far from the training views, primarily due to disocclusions, geometric uncertainty, and limited generalization of feed-forward models. When extrapolating beyond the input camera distribution, unseen regions often lead to artifacts such as blurring, ghosting, or incorrect geometry, as existing methods rely heavily on local consistency and struggle to infer plausible content for occluded areas. Additionally, view-dependent effects and complex light transport further complicate synthesis, requiring models to reason beyond interpolation-based priors. Addressing this challenge demands advancements in scene understanding, robust geometric priors, and techniques that can hallucinate missing details while maintaining consistency across novel views.

\subsection{Long Context Input}
Existing methods for 3D geometry reasoning and novel view synthesis often rely on full attention mechanisms, which lead to a cubic increase in token count and computational cost. For example, inferring from 50 images with VGGT~\cite{wang2025vggt} requires approximately 21 GB of GPU memory, while scaling to 150 images - even with advanced techniques like FlashAttention2~\cite{dao2023flashattention2} - demands around 43 GB. Training on more than 32 views remains infeasible even on the most powerful GPUs.
A promising alternative is the use of recurrent mechanisms, such as Cut3R~\cite{wang2025continuous}, which incrementally integrate new views while maintaining a memory state. Although this approach keeps inference memory usage consistently low (\ie, around 8 GB in practice), it suffers from forgetting previously seen information, leading to significant performance degradation as the number of input views increases.
Efficiently reasoning over hundreds or even thousands of views while keeping memory and computation costs manageable remains an open and pressing challenge.

\section{Social Impacts}
\label{social}
Feed-forward 3D reconstruction and view synthesis have gained considerable attention recently due to their broad applications across various industries. This section will discuss its applications and misuses from a societal aspect.

\subsection{Applications}
3D reconstruction models have a wide range of applications with positive societal impacts. To name a few, they have the potential to transform the film and gaming industries with more realistic visual effects and production speed by using reconstructed or generated 3D assets. They are also valuable in the development of smart cities, where they can be used to create "digital twins" of critical infrastructure for simulation and maintenance planning. Additionally, 3D reconstruction can help cultural heritage preservation, as it allows ancient artifacts and statues to be digitally preserved before they deteriorate.

\subsection{Misuse}
The widespread availability of 3D reconstruction models could introduce various misuses. One typical concern is related to privacy. For example, private property could be reconstructed without the owner's permission simply by taking a few pictures. To address this issue, new regulations should be established as 3D reconstruction technologies become increasingly accessible.
In addition, the generative capabilities of feed-forward 3D reconstruction models can be misused to create false evidence, such as fabricated crime scenes. To handle such misuse, advanced detection models should be developed that can distinguish between generated and real content. People can also develop techniques to add "invisible watermarks" on generated outputs, allowing simple decoding to verify if content is artificially created.

\subsection{Environment} 
Feed-forward 3D reconstruction models inherently demand substantial GPU resources and energy because they usually need to learn generic scene priors from large-scale datasets. Their inference stage, though, is more efficient: unlike optimization-based methods that update network weights at runtime, feed-forward models produce the results in a single pass within seconds. To further reduce computational costs, a promising research direction is to improve model generalizability. A pretrained model with strong generalization across diverse datasets can significantly accelerate downstream training by offering rich semantic information.
\section{Conclusion}
\label{conclusion}
Feed-forward 3D reconstruction and view synthesis have redefined the landscape of 3D vision, enabling real-time, generalizable, and scalable 3D understanding across a wide range of tasks and applications. This review covers the main approaches in feed-forward 3D reconstruction and view synthesis. Specifically, we provide an overview of these methods based on their underlying representations, such as NeRF, 3DGS, and Pointmap.
In addition, we discuss the tasks and applications of the feed-forward approaches, ranging from image and video generation to various types of 3D reconstruction. We also introduce commonly used datasets and evaluation metrics for assessing the performance of 3D feed-forward models in these tasks. Finally, we summarize the open challenges and future directions, including the need for more diverse modalities, more accurate reconstruction, free-viewpoint synthesis, and long-context generation.

\section*{Acknowledgments}

Jiahui Zhang, Muyu Xu, Kunhao Liu and Shijian Lu are funded by the Ministry of Education Singapore, under the Tier-2 project scheme with a project number MOE-T2EP20123-0003.

\bibliographystyle{alpha-doi} 
\bibliography{bibliography_short,main}

\end{document}